\documentclass[pdflatex,sn-mathphys-num]{sn-jnl}

\usepackage{graphicx}%
\usepackage{multirow}%
\usepackage{amsmath,amssymb,amsfonts}%
\usepackage{amsthm}%
\usepackage{mathrsfs}%
\usepackage[title]{appendix}%
\usepackage{xcolor}%
\usepackage{textcomp}%
\usepackage{manyfoot}%
\usepackage{booktabs}%
\usepackage{algorithm}%
\usepackage{algorithmicx}%
\usepackage{algpseudocode}%
\usepackage{listings}%
\usepackage{subfig}
\usepackage{lineno}
\usepackage{hyperref}

\theoremstyle{thmstyleone}%
\theoremstyle{thmstyletwo}%

\theoremstyle{thmstylethree}%

\begin{document}
\title[KoopGen: Koopman generator networks for representing and predicting dynamical systems with continuous spectra]{KoopGen: Koopman generator networks for representing and predicting dynamical systems with continuous spectra}


 \author[1]{\fnm{Liangyu} \sur{Su} }\email{suliangyu0917@stu.xjtu.edu.cn}
 \author[1]{\fnm{Jun} \sur{Shu} }\email{junshu@mail.xjtu.edu.cn}
 \author[2]{\fnm{Rui} \sur{Liu} }\email{scliurui@scut.edu.cn}
 \author[1]{\fnm{Deyu} \sur{Meng} }\email{dymeng@xjtu.edu.cn}
 \author[1]{\fnm{Zongben} \sur{Xu} }\email{zbxu@xjtu.edu.cn}
 \affil[1]{\orgdiv{School of Mathematics and Statistics and Ministry of Education Key Lab of Intelligent Networks and Network Security}, \orgname{Xi’an Jiaotong University}, \orgaddress{\city{Xi’an}, \postcode{710049}, \state{Shaanxi}, \country{China}}}

 \affil[2]{\orgdiv{School of Mathematics}, \orgname{South China University of Technology}, \orgaddress{\city{Guangzhou}, \postcode{510640}, \state{Guangdong}, \country{China}}}



\abstract{Representing and predicting high-dimensional and spatiotemporally chaotic dynamical systems remains a fundamental challenge in dynamical systems and machine learning. Although data-driven models can achieve accurate short-term forecasts, they often lack stability, interpretability, and scalability in regimes dominated by broadband or continuous spectra. Koopman-based approaches provide a principled linear perspective on nonlinear dynamics, but existing methods rely on restrictive finite-dimensional assumptions or explicit spectral parameterizations that degrade in high-dimensional settings. Against these issues, we introduce KoopGen, a generator-based neural Koopman framework that models dynamics through a structured, state-dependent representation of Koopman generators. By exploiting the intrinsic Cartesian decomposition into skew-adjoint and self-adjoint components, KoopGen separates conservative transport from irreversible dissipation while enforcing exact operator-theoretic constraints during learning. Across systems ranging from nonlinear oscillators to high-dimensional chaotic and spatiotemporal dynamics, KoopGen improves prediction accuracy and stability, while clarifying which components of continuous-spectrum dynamics admit interpretable and learnable representations.}




\maketitle

\section{Introduction}\label{sec1}

The rapid growth of sensing and data acquisition technologies has led to an unprecedented availability of high-resolution dynamical data, spanning turbulent flows and climate systems to neural recordings and engineered infrastructures \cite{vinuesa2022enhancing,durstewitz2023reconstructing}. While such data encode the intrinsic laws and multiscale interactions governing complex systems, extracting interpretable and predictive dynamical structure from raw time series remains a fundamental challenge, central to scientific discovery, system control, and forecasting across a wide range of disciplines \cite{brunton2022modern,wang2023scientific}.

Many data-driven approaches have been developed to model nonlinear dynamics, each emphasizing different trade-offs between interpretability, scalability, and accuracy. Traditional numerical methods, including sparse regression (e.g., SINDy~\cite{brunton2016discovering,champion2019data}) and geometric techniques such as diffusion maps and time-delay embeddings \cite{giannakis2012data,berry2013time}, recover explicit equations or low-dimensional structures when certain assumption exist, but struggle for high-dimensional and highly nonlinear systems. In contrast, modern neural network–based models, including reservoir computing and Neural ODEs \cite{pathak2018model,lukovsevivcius2009reservoir,chen2018neural,rubanova2019latent}, achieve strong short-term prediction but typically lack interpretability and exhibit limited long-term stability. Consequently, few existing approaches jointly provide interpretability, scalability, and robust long-horizon prediction.

The Koopman operator framework offers a principled alternative by lifting nonlinear dynamics to a linear evolution in an infinite-dimensional function space \cite{koopman1931hamiltonian,koopman1932dynamical}. This perspective enables spectral analysis of nonlinear systems, yielding interpretable modal structures and improved long-term stability without restrictive sparsity or low-dimensional assumptions \cite{brunton2022modern}. However, the Koopman operator is inherently infinite dimensional. Finite-mode representations are only possible for systems with discrete spectra, whereas generic nonlinear systems are characterized by continuous spectra, implying a continuum of interacting modes.

Classical data-driven methods, such as dynamic mode decomposition (DMD), approximate the Koopman operator by projection onto a finite-dimensional subspace \cite{schmid2010dynamic,williams2015data,rosenfeld2022dynamic}. Although asymptotically justified, for systems dominated by continuous spectra there exist no finite-dimensional invariant subspaces, and in practice only finite and noisy data are available. Consequently, such approximations yield truncated operators that inadequately capture the global spectral structure, thereby limiting accuracy and generalization in complex and high-dimensional systems.



To address above limitation, deep learning approaches offer a promising avenue, which have demonstrated the ability to approximate smooth mappings and effectively represent continuously varying spectral structures derived from data~\cite{brunton2024promising}. However, most deep Koopman and neural dynamical models \cite{takeishi2017learning,otto2019linearly,yeung2019learning,eivazi2021recurrent,azencot2020forecasting,alford2022deep,nayak2025temporally} continue to rely on fixed or explicitly parameterized finite-dimensional operators in a learned observation space. For systems dominated by continuous spectra, this assumption is generically violated, leading to spectral truncation, instability, and performance  degradation in chaotic and high-dimensional regimes. The DeepKoopman model proposed by Lusch et al.~\cite{lusch2018deep} demonstrated the feasibility of state-dependent operator parameterization, but challenges in scalability, stability, and generalization remain.
\begin{figure}[tb]
    \centering
    \includegraphics[width=0.95\linewidth]{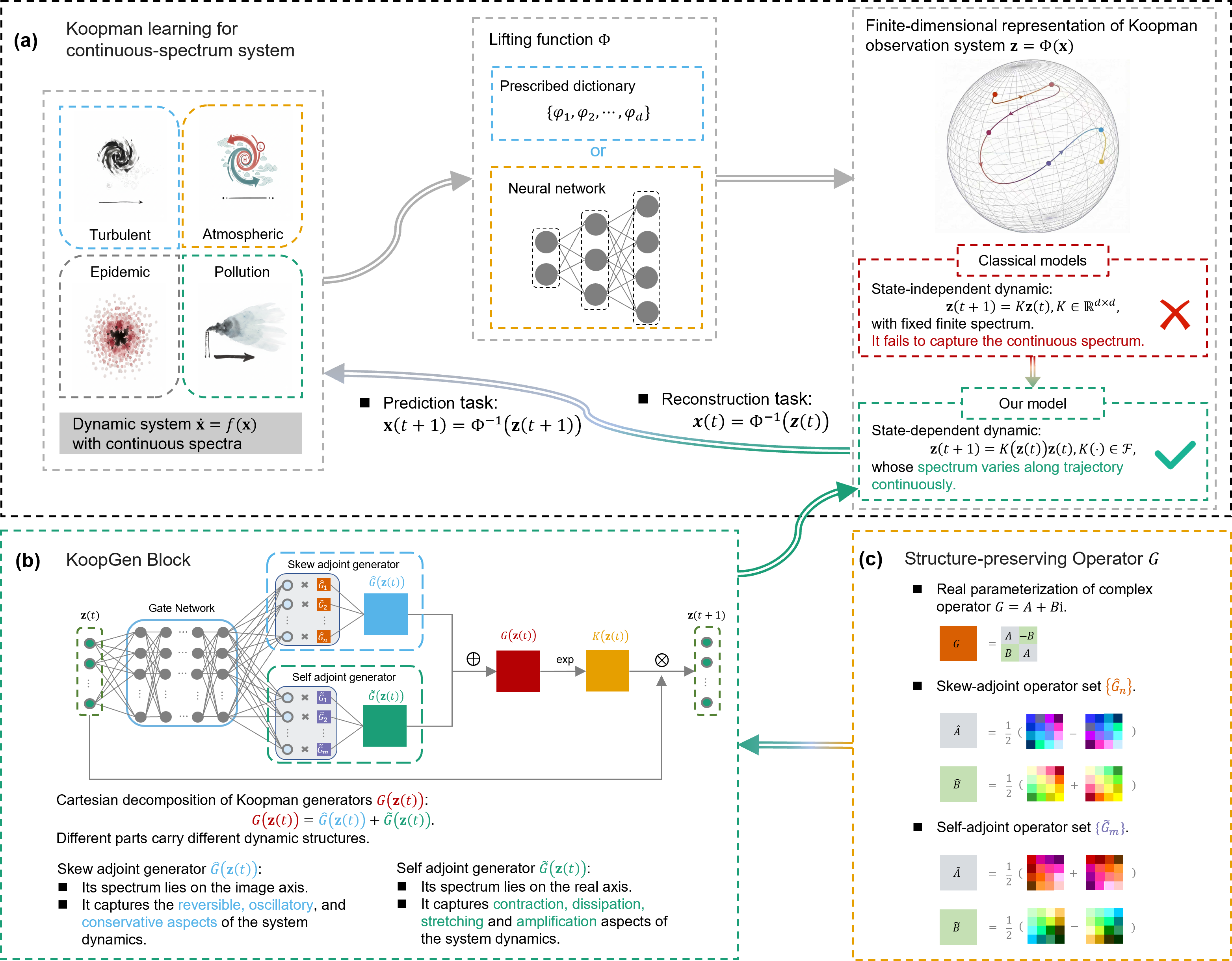}
    \caption{\textbf{Overview of the Koopman Generator model (KoopGen) for representing and predicting dynamical systems with the continuous spectrum.} \textbf{(a)} In the data-driven Koopman learning framework, system is first lifted to an observation space, $\mathbf z(t)=\Phi(\mathbf x(t))$, where the dynamics are modeled linearly. The classical \textit{state-independent} model is unable to capture the complex behavior of continuous-spectrum systems. KoopGen instead employ a \textit{state-dependent} formulation, allowing the operator spectrum to vary continuously along trajectories and enabling accurate representation of complex dynamics. \textbf{(b)} The KoopGen model consists of a gating network and two operator sets, $\{\widehat G^n\}_{n=1}^N$ and $\{\widetilde G^m\}_{m=1}^M$, corresponding to skew-adjoint (conservative) and self-adjoint (dissipative) components, respectively. The gating network adaptively weights these generators based on the current latent state $\mathbf z(t)$, yielding a state-dependent Koopman operator $K(\mathbf z(t))$ that advances the dynamics forward in time. \textbf{(c)} Each generator is parameterized in a structure-preserving real block form, ensuring exact enforcement of the skew-adjoint and self-adjoint constraints and encoding physically interpretable dynamical properties.}
    \label{fig:koopgen}
\end{figure}

Here, we introduce Koopman Generator model (KoopGen), a mathematically grounded, physically interpretable, and data-adaptive neural architecture (as shown in Fig.~\ref{fig:koopgen}) for learning dynamical systems with continuous spectra. KoopGen embeds the Cartesian decomposition of the Koopman generator directly into the model design, enabling a principled separation of conservative and dissipative dynamics within a state-dependent operator framework. This structure enhances interpretability and long-term stability, while  scaling to high-dimensional and spatiotemporally chaotic systems in regimes that are beyond the reach of existing approaches. 

 
Most notably, our work introduce the first unified framework capable of accurately representing and predicting systems across a broad spectrum of complexity, from simple nonlinear oscillators to strongly chaotic attractors, and from low-dimensional dynamics to high-dimensional spatiotemporal chaos. Through extensive experiments on four representative continuous-spectrum systems, including a nonlinear pendulum, the low-dimensional chaotic Lorenz-63,  the high-dimensional chaotic Lorenz-96 and the spatiotemporally chaotic Kuramoto–Sivashinsky equation, we demonstrate that our approach not only surpasses existing baselines in long-term predictive accuracy and representation quality, but also achieves a substantial gains in scalability, interpretability, and physical fidelity. These results establish KoopGen as a new operator-theoretic framework for learning complex system, resolving a longstanding limitation that has hindered progress in Koopman theory and application.

\section{Results}\label{sec2}

\subsection{Koopman operator, generator and spectral decomposition}

Koopman operator theory provides a linear perspective on nonlinear dynamical systems. 
For a system $\dot{\mathbf x}=f(\mathbf x)$ with flow map $F^t$ and a complex Hilbert space of observables $\mathcal F$, the Koopman operator $\mathcal K^t:\mathcal F\to\mathcal F$ is defined by
\begin{equation}
    \mathcal K^t g = g \circ F^t, \ \ \ g\in \mathcal F.
\end{equation}
Although the state dynamics $F^t$ may be nonlinear, $\mathcal K^t$ is linear on $\mathcal F$. 
The associated generator $\mathcal G$ of the Koopman semigroup $\{\mathcal K^t\}_{t\geq 0}$ is defined as 
\begin{equation}
    \mathcal G g = \lim_{t\to 0^+} \frac{\mathcal K^t g - g}{t}.
\end{equation}
By the spectral theorem~\cite{van2022functional}, under some mild conditions, Koopman operator $\mathcal K^t$ can be decomposed as
\begin{equation}
    \mathcal K^t = \sum_{\lambda\in\sigma_p} e^{t\lambda}\mathbb P_\lambda +\int_{\sigma_c}e^{t\lambda}d\mathbb E_c(\lambda), 
\end{equation}
where $\mathbb P_\lambda$ denotes the spectral projection onto the eigenspace for an isolated eigenvalue $\lambda\in \sigma_p$ and $\mathbb E_c(\cdot)$ is the projection-valued measure of the continuous spectrum $\sigma_c$. 

When the Koopman spectrum is purely discrete, each projection $\mathbb P_\lambda$ reduces to an expansion in Koopman eigenfunction $\phi_\lambda$,
\begin{equation}
    \mathbb P_\lambda g = \langle g, \phi_\lambda \rangle \phi_\lambda.
\end{equation}
In this case, observables in the span of eigenfunctions admit the representation
\begin{equation}
    g = \sum_{\lambda\in\sigma_p} a_\lambda \phi_\lambda,
    \qquad 
    \mathcal K^t g = \sum_{\lambda\in\sigma_p} a_\lambda e^{t\lambda} \phi_\lambda,
\end{equation}
providing a complete and explicit linear characterization of their evolution.

However, many physically relevant systems possess a continuous spectrum. In such settings, the Koopman eigenfunction may not exist or may fail to span any informative subspace, making spectral representations substantially more difficult and highlighting the need for alternative learning-based formulations.

\subsection{Data-driven Koopman learning}

In practice, any approximation of the Koopman operator implicitly relies on two fundamental components: a observation space $\text{Span}\{\varphi_1,\varphi_2,\cdots,\varphi_d\}$, where $\varphi_i$ is a scalar-valued function, and a finite dimensional representation $K$ of the Koopman operator restricted to this space. Abstractly, this can be expressed as
\begin{equation}\label{eq:dmd}
    \Phi\left(F(\mathbf x(t))\right) = K\mathbf z(t), \qquad  \mathbf z(t) = \Phi\left(\mathbf x(t)\right)\in\mathbb C^d,
\end{equation}
where $\Phi=\left(\varphi_1,\varphi_2,\cdots,\varphi_d\right)$ is a vector-valued observation function.

Classical methods such as DMD~\cite{schmid2010dynamic} and extended DMD (EDMD)~\cite{williams2015data} rely on a prescribed dictionary of observables. More recent neural Koopman models, such as LRAN and its variants~\cite{otto2019linearly, yeung2019learning, azencot2020forecasting, alford2022deep, nayak2025temporally}, replace this fixed dictionary with learned observation functions. Despite their algorithmic differences, all these methods share the same underlying assumption: the existence of a finite-dimensional subspace that is invariant under the Koopman operator.
Such an assumption is consistent with systems whose Koopman spectrum is discrete, for which eigenfunctions provide a natural basis. However, for systems with a significant continuous spectrum, nontrivial finite-dimensional invariant subspaces generally do not exist.

This limitation motivates relaxing the fixed Koopman operator $K$ in Eq.~\ref{eq:dmd} to a state-dependent operator $K(\mathbf z(t))$, allowing the operator spectrum to vary continuously along the trajectories,
\begin{equation}\label{eq:framework}
\Phi\left(F(\mathbf x(t))\right) = K(\mathbf z(t))\mathbf z(t),
\qquad
\mathbf z(t) = \Phi\left(\mathbf x(t)\right) \in \mathbb C^d .
\end{equation}
Early efforts of this idea is the DeepKoopman model proposed by Lusch et al.~\cite{lusch2018deep}, which could interpreted as learning both the $\Phi$ and the operator $K(\mathbf z(t))$ whose structure is constrained to be diagonal with spectra $\{\lambda_i\}_{i=1}^k$ predicted by an auxiliary neural network.

\begin{figure}[tb]
    \centering
    \subfloat[\label{fig:nrmse_pen}]{\includegraphics[width=0.45\linewidth]{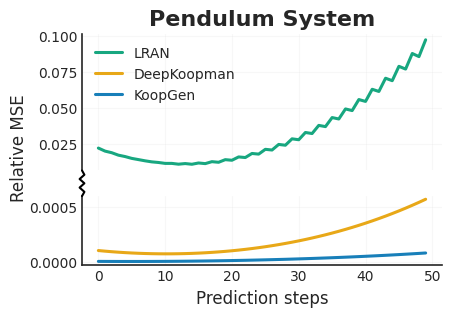}}
    \hfill
    \subfloat[\label{fig:nrmse_63}]{\includegraphics[width=0.45\linewidth]{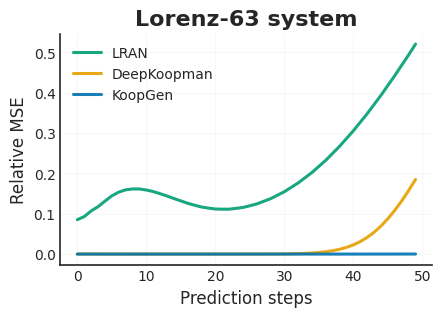}}
    \hfill
    \subfloat[\label{fig:nrmse_96}]{\includegraphics[width=0.45\linewidth]{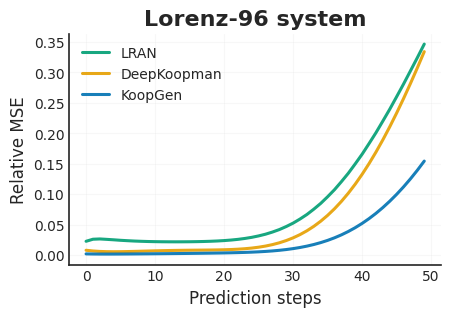}}
    \hfill
    \subfloat[\label{fig:nrmse_ks}]{\includegraphics[width=0.45\linewidth]{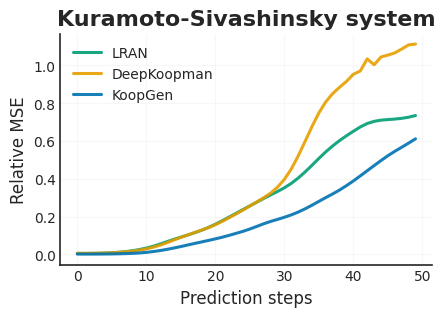}}
    \caption{Prediction results comparison for KoopGen, DeepKoopman and LRAN across different system. The horizontal axis denotes the prediction horizon, and the vertical axis reports the prediction error. LRAN~\cite{otto2019linearly} represents Koopman models with a state-independent transfer operator (Eq.~\ref{eq:dmd}), whereas DeepKoopman~\cite{lusch2018deep} corresponds to the state-of-the-art state-dependent Koopman approach (Eq.~\ref{eq:framework}). The first row shows results for two low-dimensional continuous-spectrum systems, while the second row presents two high-dimensional continuous-spectrum systems. DeepKoopman improves substantially over state-independent methods in low-dimensional settings but degrades markedly in high dimensions, indicating limited scalability. In contrast, KoopGen consistently achieves the lowest prediction errors across all systems, demonstrating superior accuracy, stability, and scalability for continuous-spectrum dynamics.}
    \label{fig:nrmse}
\end{figure}

The empirical success of DeepKoopman on typical low-dimensional continuous-spectrum systems, such as the nonlinear pendulum and the Lorenz–63 system (Fig.~\ref{fig:nrmse_pen} and Fig.~\ref{fig:nrmse_63}), marks an important breakthrough for Koopman-based learning under continuous spectra. Nevertheless, this success does not extend to high-dimensional dynamics. As shown in Fig.~\ref{fig:nrmse_96} and Fig.~\ref{fig:nrmse_ks}, DeepKoopman exhibits pronounced performance degradation on the Lorenz–96 and Kuramoto–Sivashinsky systems, highlighting fundamental challenges in scaling to high-dimensional chaotic regimes. These limitations stem from the explicit parameterization of the spectrum. First, the model does not explicitly encode physical or dynamical structure, which hampers long-term generalization when complex, multiscale interactions dominate the dynamics. Second, the number of predicted spectra grows with system dimensionality, leading to poor scalability and increased computational burden. Finally, direct spectral modeling is prone to numerical instabilities during training and long-horizon prediction, reducing robustness in strongly chaotic or spatiotemporally complex systems.

\subsection{A Koopman generator structure in continuous-spectrum dynamics}\label{sec:koopgen}

The main goal of this work is to design an interpretable, generalizable and scalable model capable of representing and predicting various dynamical systems with continuous spectra. To this end, we introduce the Koopman Generator model (KoopGen), a Koopman operator learning framework that parameterizes infinitesimal generators of dynamics, which encode differential structure of the flow and admit well-defined operator-theoretic decompositions. Modeling the generator, rather than the Koopman operator itself, enables an interpretable separation of conservative and dissipative mechanisms and provides a structured pathway to construct stable discrete-time operators via exponentiation. We begin by reviewing the Cartesian decomposition of Koopman generators, which underpins the design of our model.

\subsubsection{Cartesian decomposition of Koopman generators}

Since the Koopman operator and its generator are linear operators acting on complex-valued Hilbert spaces, their structure admits a Cartesian decomposition, analogous to separating a complex number into its real and imaginary parts. Specifically, for a bounded Koopman generator $\mathcal G$, a unique Cartesian decomposition exists \cite{van2022functional}
\begin{equation}
    \mathcal G=\hat {\mathcal G} + \tilde {\mathcal G}, \qquad \hat {\mathcal G}^\ast=-\hat {\mathcal G}, \qquad \tilde {\mathcal G}^\ast=\tilde {\mathcal G},
\end{equation}
where $\widehat {\mathcal G}$ is skew-adjoint and $\widetilde {\mathcal G}$ is self-adjoint. This decomposition is not only mathematically fundamental, but also carries clear dynamical significance.

Based on the Stone’s theorem \cite{stone1932one}, the skew-adjoint generator $\widehat {\mathcal G}$ generates a unitary Koopman semigroup $\{\widehat {\mathcal K}^{t}\}_{t>0}$. Unitary evolution preserves inner products and norms, implying that $\widehat {\mathcal G}$ captures the reversible, oscillatory, and conservative aspects of the system dynamics. In particular, such skew-adjoint dynamics do not create or dissipate energy, but rather redistribute it across modes, which is a hallmark of conservative transport processes. This behavior typically arises from rotational flows and Hamiltonian subsystems, where the total energy is preserved while being transferred among interacting components, and no contraction or expansion occurs.

In contrast, the self-adjoint part $\widetilde {\mathcal G}$ generates a positive semigroup $\{\widetilde {\mathcal K}^{t}\}_{t>0}$, whose spectral properties are determined by the real spectrum of $\widetilde {\mathcal G}$. Negative spectral components correspond to contraction and dissipation, leading to attractor formation and energy decay, while positive components encode stretching and amplification, as commonly observed in chaotic and hyperbolic dynamics. 

This decomposition suggests a natural design principle for data-driven Koopman models. Rather than learning a generic generator, explicitly parameterizing its skew-adjoint and self-adjoint components preserve their distinct structural and dynamical roles. Such a structure-preserving parameterization ensures that conservative transport and irreversible processes are modeled in a controlled and interpretable manner, while remaining compatible with the underlying operator theory.

\subsubsection{KoopGen model}
Guided by the Cartesian decomposition, we introduce the KoopGen model, which provides an interpretable and scalable parameterization of the Koopman generator while explicitly preserving its skew-adjoint and self-adjoint structure. Specifically, we model the generator as a state-dependent convex combination of finite learnable operators. Specifically, the skew-adjoint and self-adjoint parts are parameterized independently as
\begin{equation}
    \widehat G(\mathbf z(t)) = 
        \sum_{n=1}^{N} 
        \widehat w_n(\mathbf z(t)) \widehat G_n,
    \qquad 
    \sum_{n=1}^{N} \widehat w_n = 1,\qquad  \widehat w_n\ge 0,
\end{equation}
\begin{equation}
    \widetilde G(\mathbf z(t)) = 
        \sum_{m=1}^{M} 
        \widetilde w_m(\mathbf z(t)) \widetilde G_m,
    \qquad 
    \sum_{m=1}^{M} \widetilde w_m = 1,\qquad  \widetilde w_m\ge 0,
\end{equation}
where $\{\widehat G_n\}_{n=1}^N$ and $\{\widetilde G_m\}_{m=1}^M$ are sets of learnable skew-adjoint and self-adjoint operators, respectively, and the gating networks $\widehat w(\cdot)$ and $\widetilde w(\cdot)$ adaptively weight their contributions based on the state. 

The above construction is mathematically well grounded. Both the self-adjoint and skew-adjoint operator classes are closed under convex combination, ensuring that the Cartesian structure remains exact. In addition, the formulation effectively decouples the state dimension from the network parameter, thereby guaranteeing its scalability to high-dimensional systems that would be intractable for previous methods.

Finally, the learnable Koopman generator is then given by
\begin{equation}
    G(\mathbf z(t)) = 
    \widehat G(\mathbf z(t)) +
    \widetilde G(\mathbf z(t)),
\end{equation}
enabling the model to adaptively blend conservative transport and irreversible dissipation. Together, these features unify dimension-independent scalability, structural stability, and interpretable dynamics within a single framework for learning complex systems.

Based on the classical Hille–Yosida generation theorem~\cite{engel2000one}, a well-defined generator induces the Koopman operators through the exponential map, providing a principled link between continuous and discrete time dynamics. Accordingly, the discrete-time learnable Koopman operator for time step $\Delta t$ is computed by
\begin{equation}
    K^{\Delta t}(\mathbf z(t)) = 
    \exp\bigl(\Delta t\, G(\mathbf z(t))\bigr).
\end{equation}
In this work, we focus on datasets with uniformly sampled time points. Consequently, the time step $\Delta t$ is a fixed constant, and for notational simplicity, we omit the superscript in what follows, writing $K(\mathbf z(t))$ directly. Finally, the learned Koopman operator is used to advance the state forward,
\begin{equation}
    \mathbf z(t+1) = K\left(\mathbf z(t)\right)\mathbf z(t).
\end{equation}

\subsubsection{Structure-preserving real parameterizations of generators}
To ensure exact adherence to skew-adjoint and self-adjoint conditions, each learnable generator is parameterized using its canonical real block representation. In details, a learnable skew-adjoint operator is represented in complex form as
\begin{equation}
    \widehat G = \widehat A + \mathrm i \widehat B,
\qquad 
\widehat A^\top = -\widehat A,\ \qquad \widehat B^\top = \widehat B,
\end{equation}
and realized in real coordinates through
\begin{equation}
\widehat G =
\begin{bmatrix}
    \widehat A & -\widehat B \\
    \widehat B & \widehat A
\end{bmatrix},
\qquad 
\widehat A = \frac12(P - P^\top),\qquad 
\widehat B = \frac12(Q + Q^\top),
\end{equation}
with learnable $P,\ Q\in\mathbb R^{D\times D}$.  

Similarly, each learnable self-adjoint takes the complex form
\begin{equation}
    \widetilde G = \widetilde A + \mathrm i \widetilde B,
\qquad 
\widetilde A^\top = \widetilde A,\ \qquad \widetilde B^\top = -\widetilde B,
\end{equation}
with real block representation
\begin{equation}
\widetilde G =
\begin{bmatrix}
    \widetilde A & -\widetilde B \\
    \widetilde B & \widetilde A
\end{bmatrix},\qquad 
\widetilde A = \frac12(U + U^\top),\qquad 
\widetilde B = \frac12(V - V^\top),
\end{equation}
with learnable $U,\ V\in\mathbb R^{D\times D}$.  

This parameterization guarantees exact structural preservation while remaining fully differentiable and efficient, enabling stable training of model that respect both mathematical constraints and physical consistency.

\subsection{Numerical Examples}

We evaluate the proposed KoopGen framework on four representative dynamical systems spanning increasing nonlinearity, chaos, and dimensionality. This hierarchy enables a systematic assessment of representation accuracy, scalability, and long-term predictive stability in continuous-spectrum regimes.

KoopGen is compared with two representative Koopman-based neural models: LRAN \cite{otto2019linearly}, which assumes a state-independent Koopman operator (Eq.~\ref{eq:dmd}), and DeepKoopman \cite{lusch2018deep}, which allows state-dependent operators via explicit spectral parameterization (Eq.~\ref{eq:framework}). As summarized in Fig.~\ref{fig:nrmse}, DeepKoopman consistently improves upon LRAN in low-dimensional systems, confirming the benefit of relaxing the fixed-operator assumption when finite-dimensional invariant subspaces do not exist. However, this advantage degrades rapidly with increasing dimensionality: in high-dimensional chaotic systems, DeepKoopman exhibits accelerated error growth and reduced robustness. In contrast, KoopGen achieves uniformly lower prediction errors across all systems, with the performance gap widening at longer horizons. These results demonstrate that KoopGen provides both improved short-term accuracy and markedly enhanced long-term stability, particularly in high-dimensional continuous-spectrum systems.

\subsubsection{Nonlinear pendulum}
\begin{figure}[tb]
    \centering
    \subfloat[\label{fig:pen_eigen}]{\includegraphics[height=5cm]{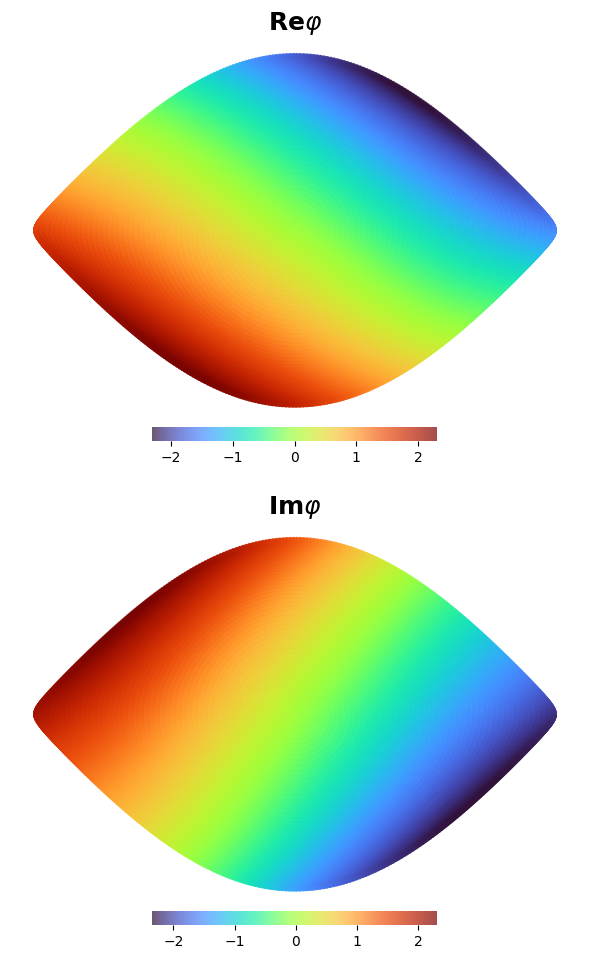}}
    \hfill
    \subfloat[\label{fig:pen_mag}]{\includegraphics[height=5cm]{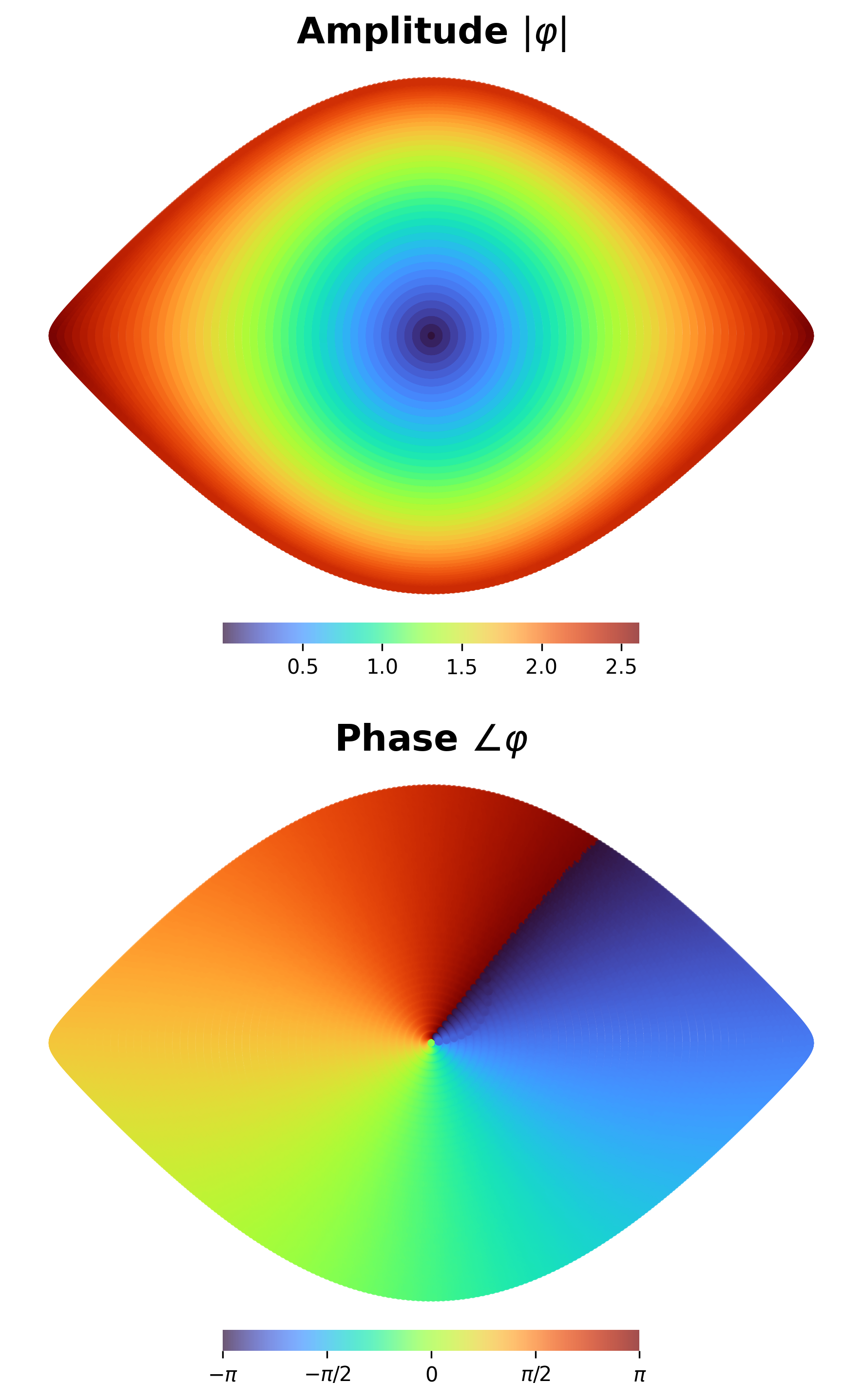}}
    \hfill
    \subfloat[\label{fig:pen_weight}]{\includegraphics[height=5cm]{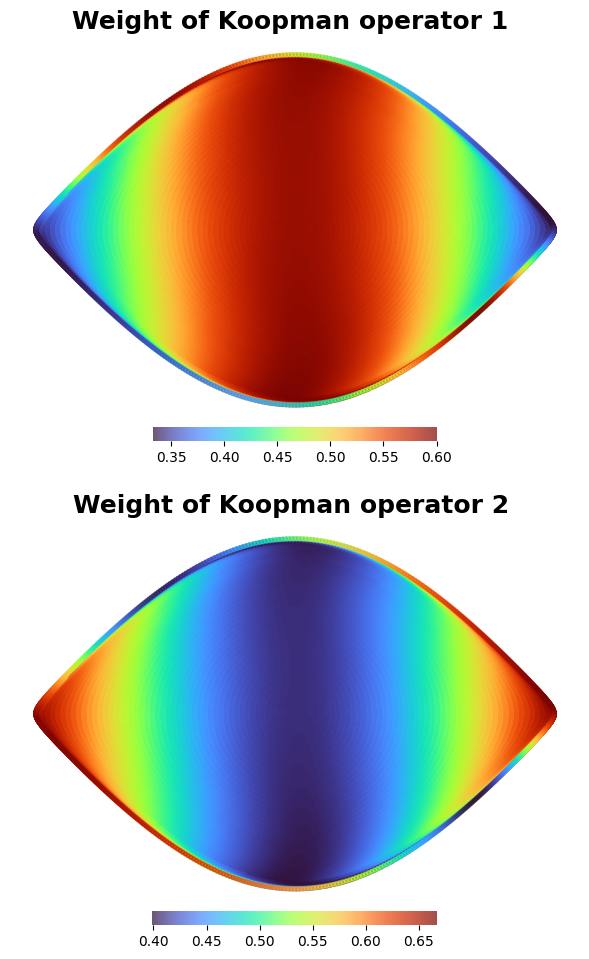}}
    \hfill
    \subfloat[\label{fig:pen_}]{\includegraphics[height=5cm]{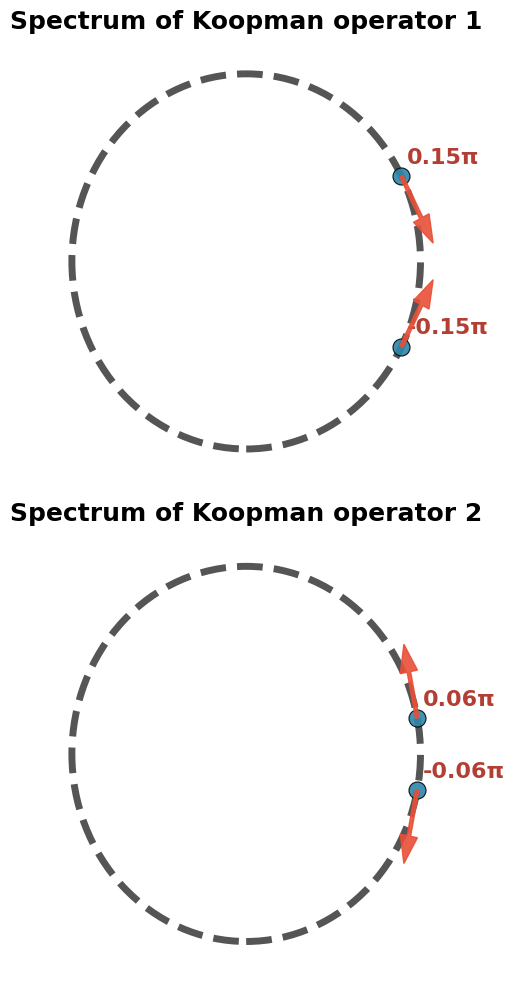}}
    \caption{Learned Koopman eigenfunction and operator structure for the nonlinear pendulum. The eigenfunction exhibits a magnitude aligned with Hamiltonian energy level sets and a smoothly varying phase that parameterizes the periodic motion. The learned Koopman spectra reveal two operators with eigenvalues on the unit circle at angular frequencies $\pm0.15\pi$ and $\pm0.06\pi$, corresponding to counterclockwise and clockwise rotational dynamics. The spatial distribution of operator weights shows clear physical organization: the counterclockwise mode dominates near turning points, while the clockwise mode is activated near equilibrium crossings. Together, these operators provide a continuous, bidirectional decomposition of the pendulum flow, with the gating network adapting smoothly along the trajectory.}
    \label{fig:pen_eigen_pen}
\end{figure}
We first consider a nonlinear pendulum, a simple yet canonical continuous-spectrum system,
\begin{equation}
    \ddot{x}=-\sin(x), \ \ \ \Rightarrow\ \ \ \left\{\begin{aligned}
        &\dot{x}_1=x_2,\\ &\dot{x}_2=-\sin(x_1).
    \end{aligned}\right. 
\end{equation}
Despite its simplicity, the system exhibits energy-dependent frequencies that challenge most neural methods. KoopGen captures this behavior through a compact eigenfunction whose magnitude aligns with Hamiltonian energy level sets and whose phase varies smoothly along each orbit, effectively parameterizing the periodic motion and providing an intrinsic phase coordinate that tracks the dynamical progression along each orbit (Fig.~\ref{fig:pen_eigen} and \ref{fig:pen_mag}).

Fig.~\ref{fig:pen_} shows the spectrum of learned unitary koopman operators, two distinct operators emerge with spectra located on the unit circle at angular frequencies of approximately $\pm0.15\pi$ and $\pm 0.06\pi$. These correspond to rotational dynamics with opposite directions of phase evolution, one counterclockwise and one clockwise, reflecting the bidirectional nature of the pendulum’s motion in phase space. The spatial distribution of operator weights in Fig.~\ref{fig:pen_weight} reveals clear physical meaning. The counterclockwise operator dominates near the turning points, where the pendulum momentarily stops and local phase evolution slows, while the clockwise operator becomes active near the equilibrium crossing, where the angular velocity is highest and the phase advances rapidly. Together, the two operators form a smooth, complementary decomposition: as the pendulum swings back and forth, the gating network continuously shifts attention between them, capturing alternating rotational modes without discontinuity.


\subsubsection{Lorenz-63 system}

We next examine the Lorenz–63 system, a prototypical low-dimensional chaotic flow,
\begin{equation}
    \left\{\begin{aligned}
        &\dot{x} = \sigma(y-x),\\ &\dot{y}=x(\rho-z)-y,\\ &\dot{z}=xy-\beta z,
    \end{aligned}\right.
\end{equation}
with $\sigma = 10, \beta = 8/3$, and $\rho = 28$, corresponding to the original system studied by Lorenz. The system arises from a truncated Galerkin approximation of the Navier–Stokes equations for thermal fluid motion, where the variables $x,y,z$ can be interpreted as the convective intensity, horizontal temperature variation, and vertical temperature variation, respectively.

As shown in Fig.~\ref{fig:nrmse_63} and Fig~\ref{fig:lor_error} (depicted in Appendix B), KoopGen achieves substantially improved prediction accuracy compared with existing methods, particularly at longer prediction horizons. This stems from KoopGen's ability to capture the intrinsic structure of the Lorenz dynamics.

\begin{figure}
    \centering
    \includegraphics[width=0.9\linewidth]{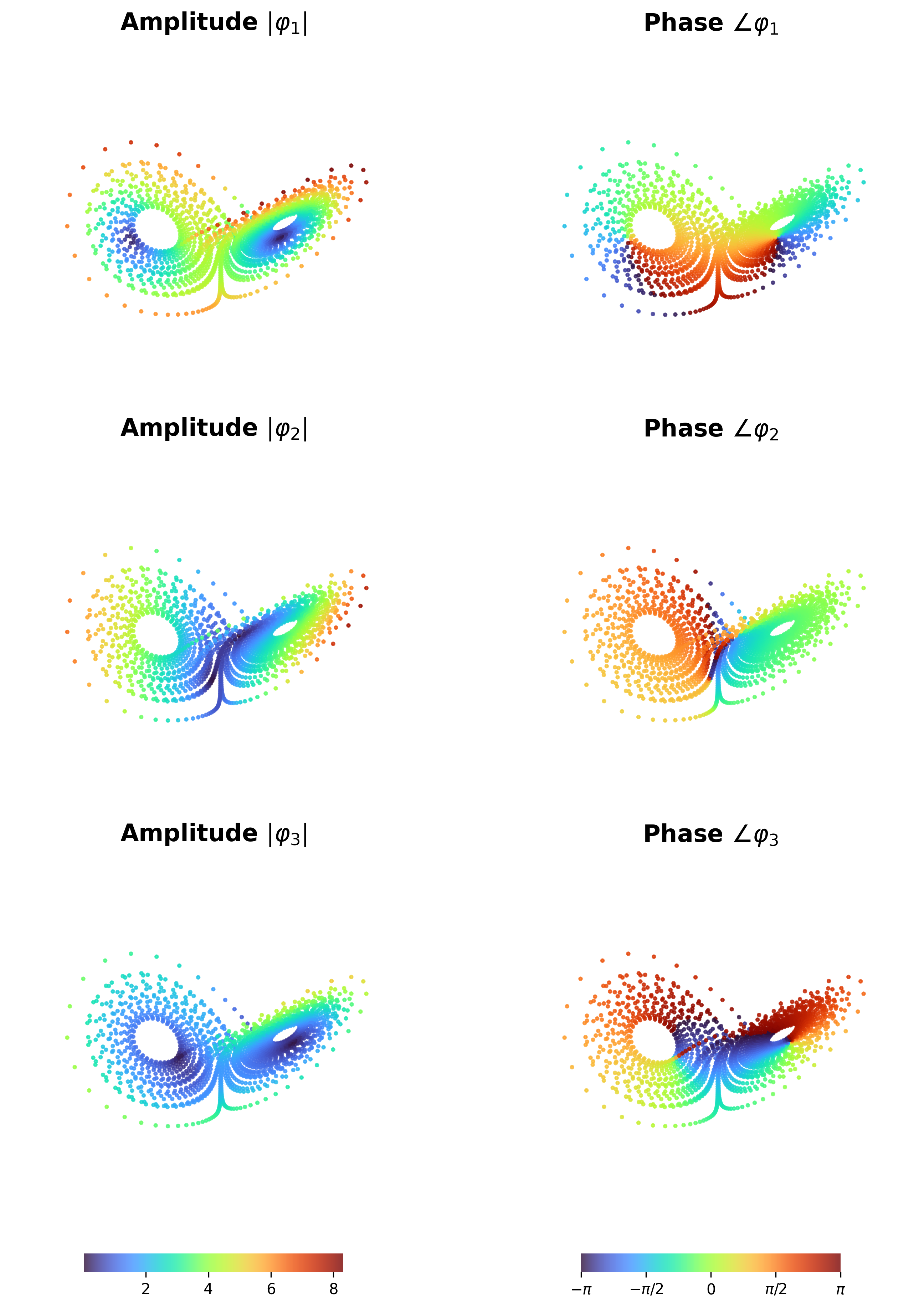}
    \caption{Illustration of magnitude and phase of the KoopGen eigenfunctions for the Lorenz-63 system. Learned Koopman eigenfunctions of the Lorenz attractor. The three eigenfunctions yield a compact and interpretable decomposition of the Lorenz dynamics. The first eigenfunction encodes the cyclic motion around each wing through a smooth phase progression and amplitude growth along the spiral arms, with large amplitudes concentrated near the inter-wing transition channel. The second eigenfunction captures the left–right symmetry of the attractor, exhibiting weak phase variation but a sign change that separates the two metastable lobes. The third eigenfunction provides a coherent phase coordinate with a continuous $2\pi$ winding across the attractor, remaining smooth even through wing-switching events. Together, the learned eigenfunctions recover the core stretching–folding mechanism of the Lorenz flow and provide an interpretable spectral representation of its chaotic dynamics.}
    \label{fig:eigen_lor}
\end{figure}

The three learned eigenfunctions provide a compact and interpretable decomposition of the Lorenz attractor. The first eigenfunction shows a smooth phase progression from $ -\pi$ to $\pi$ together with a monotonic amplitude increase along each spiral arm. This structure reflects the cyclic motion around the two wings and the gradual growth of stretching energy as trajectories move away from the core of each wing. Large amplitudes appear near the inter-wing transition channel, consistent with the higher energy required for wing switching.
The second eigenfunction exhibits a symmetric amplitude pattern across the attractor and only weak phase variation concentrated around zero. Rather than encoding rotational behavior, this mode primarily distinguishes the two wings through a sign change, reflecting the intrinsic left–right symmetry of the attractor. Its behavior is consistent with the global partitioning of the state space into two metastable regions.
The third eigenfunction shows the opposite pattern, minimal amplitude modulation but a clear and continuous $2\pi$ phase winding on attractor. Crucially, its phase remains continuous even across the wing-switching region, where trajectories undergo the rapid divergence that drives chaotic behavior. This mode therefore provides a coherent phase coordinate that tracks the rotational progression around each equilibrium and seamlessly bridges transitions between the lobes. Such continuous phase evolution across switching is a hallmark of the Lorenz flow’s stretching-and-folding mechanism, and its accurate reconstruction demonstrates that the learned Koopman representation captures the core dynamical structure underlying Lorenz chaos.

Together, the three modes separate energy growth, lobe identification, and global rotational progression, showing that the model reconstructs not only the geometry but also the essential switching and instability structures that define the chaotic dynamics of the Lorenz system. By capturing these global intrinsic dynamical structures, our approach achieves a faithful linearization of the flow in an appropriate observable space. This structural fidelity is precisely what enables stable long-term prediction in a system as shown in Fig.~\ref{fig:nrmse}.

\subsubsection{Lorenz-96 system}

\begin{figure}[tb]
    \centering
    \subfloat[\label{fig:lorenz96_1}]{\includegraphics[width=0.45\linewidth]{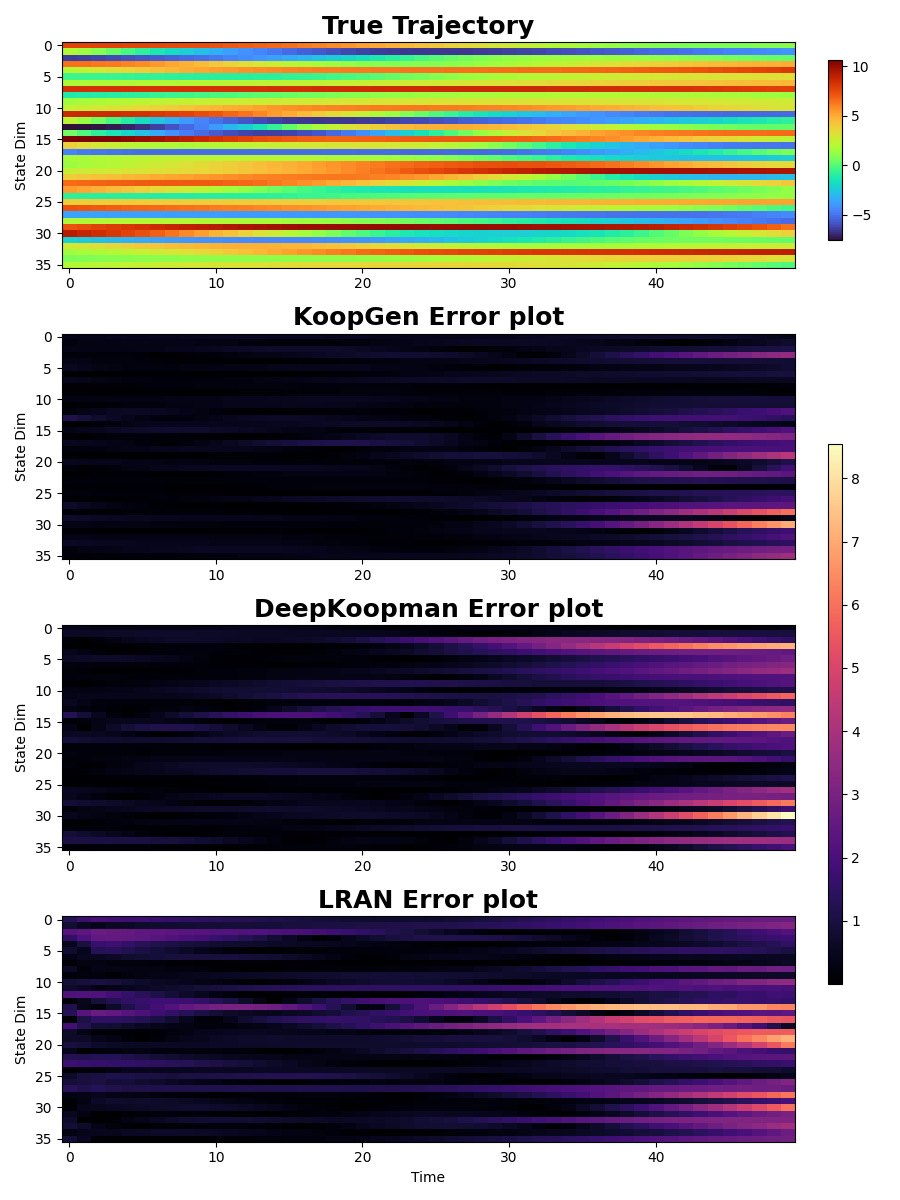}}
    \hfill
    \subfloat[\label{fig:lorenz96_2}]{\includegraphics[width=0.45\linewidth]{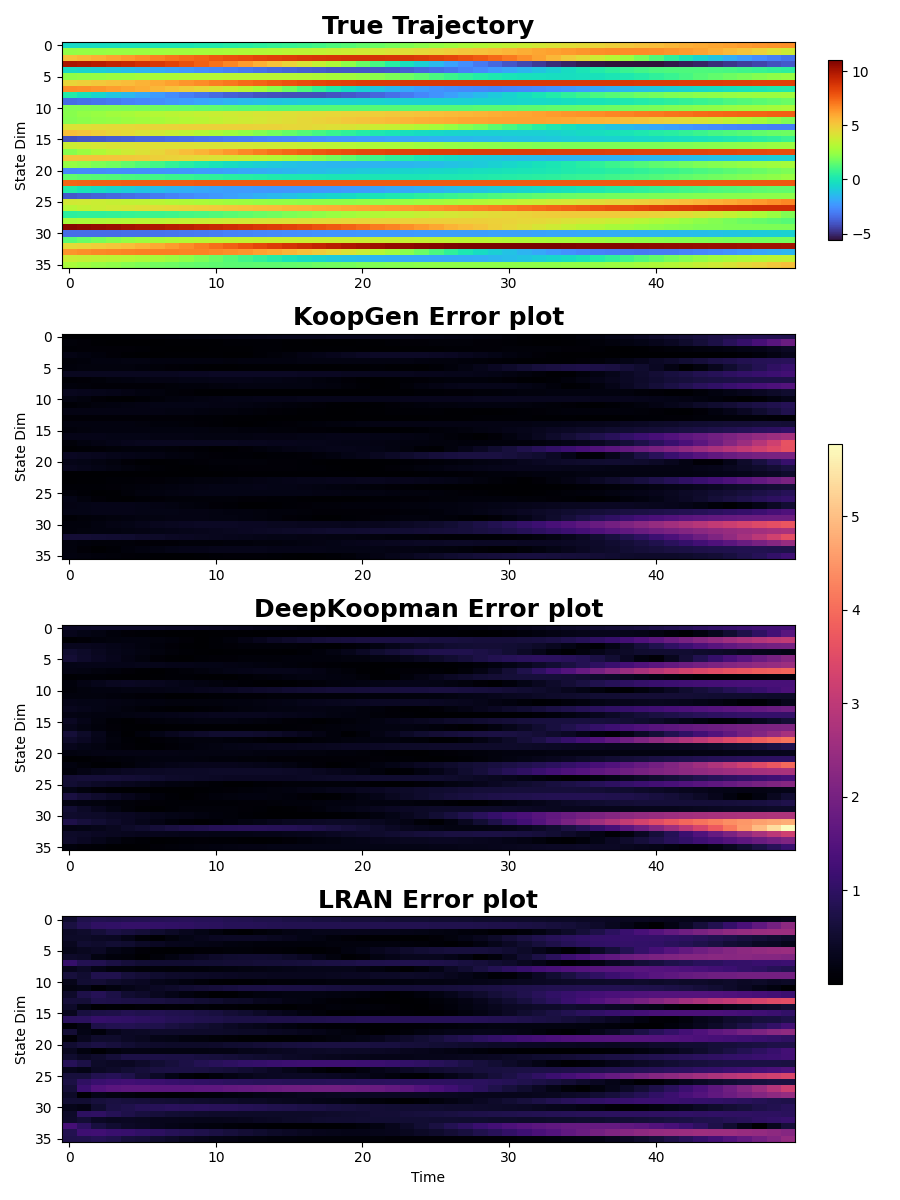}}

    \caption{Comparison of prediction errors for KoopGen, DeepKoopman and LRAN on the Lorenz–96 system.  These results illustrate that KoopGen maintains low and spatially coherent errors, whereas others exhibit faster error accumulation and loss of dynamical structure at longer times.}
    \label{fig:96_error_plot}
\end{figure}

To assess scalability, we consider the Lorenz–96 system with $K=36$ variables, a standard high-dimensional chaotic benchmark.

\begin{equation}
    \frac{dx_i}{dt}
    = (x_{i+1} - x_{i-2})\, x_{i-1} - x_i + F,
    \qquad i = 0,\ldots,K-1,
\end{equation}
with cyclic indexing imposed through $x_{i+K} = x_i$. We take $K = 36$ variables and the classical forcing parameter $F = 8$, a regime known to produce sustained high-dimensional chaotic dynamics. Each variable $x_i$ represents a scalar quantity (such as temperature or vorticity) distributed along a periodic latitude circle, and the coupling structure models advection, damping, and external forcing.

Representative examples of ground-truth trajectories and corresponding prediction errors are shown in Fig. ~\ref{fig:96_error_plot}, with additional examples provided in Appendix~\ref{apd:res}. These examples show that both the KoopGen and the DeepKoopman make quantitatively accurate short term predictions. While the predictions after $20$ steps, KoopGen outperforms others significantly as shown in Fig.~\ref{fig:nrmse_96}, demonstrating that KoopGen has a stronger representation ability and a more stable long-term prediction capability for complex high dimensional systems. 

\subsubsection{Kuramoto-Sivashinsky system}
\begin{figure}[tb]
    \centering
    \subfloat[\label{fig:KS_1}]{\includegraphics[width=0.45\linewidth]{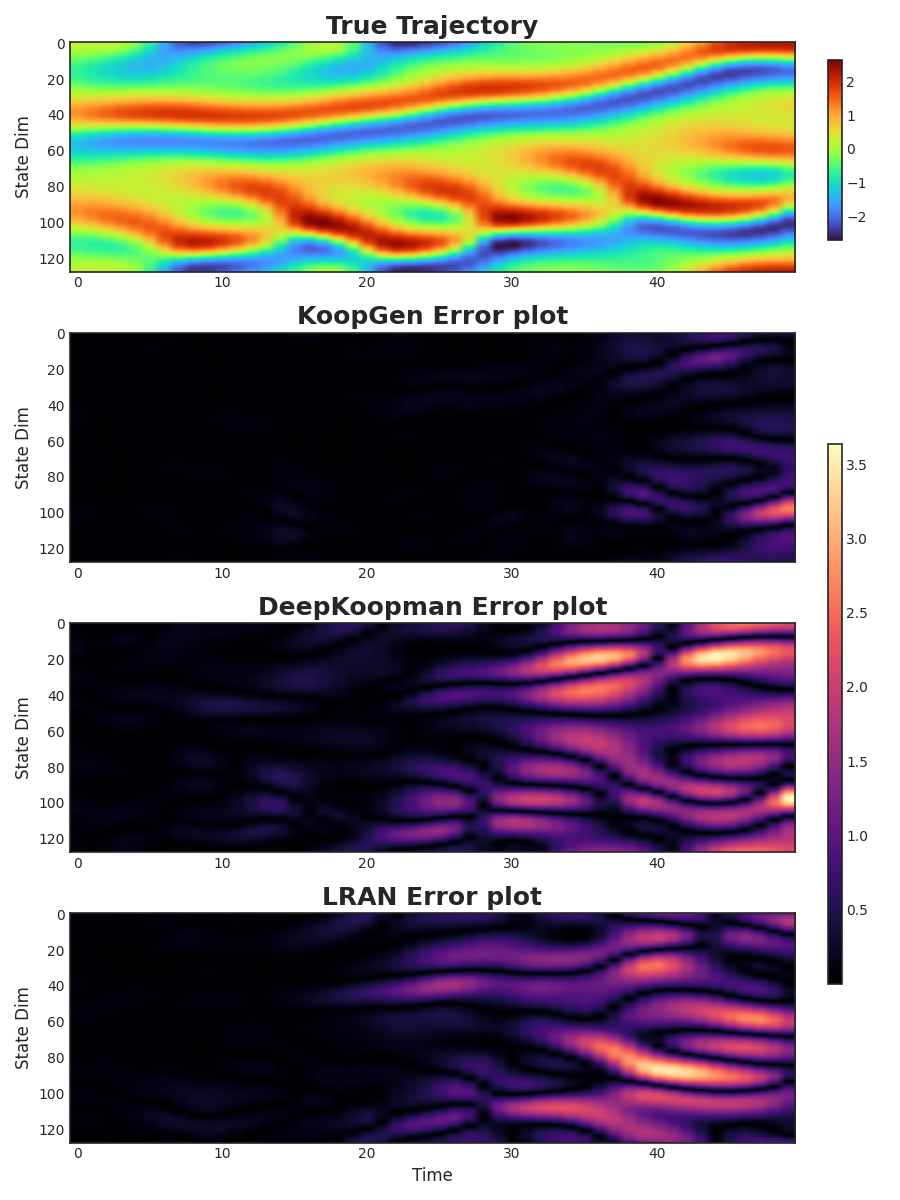}}
    \hfill
    \subfloat[\label{fig:KS_2}]{\includegraphics[width=0.45\linewidth]{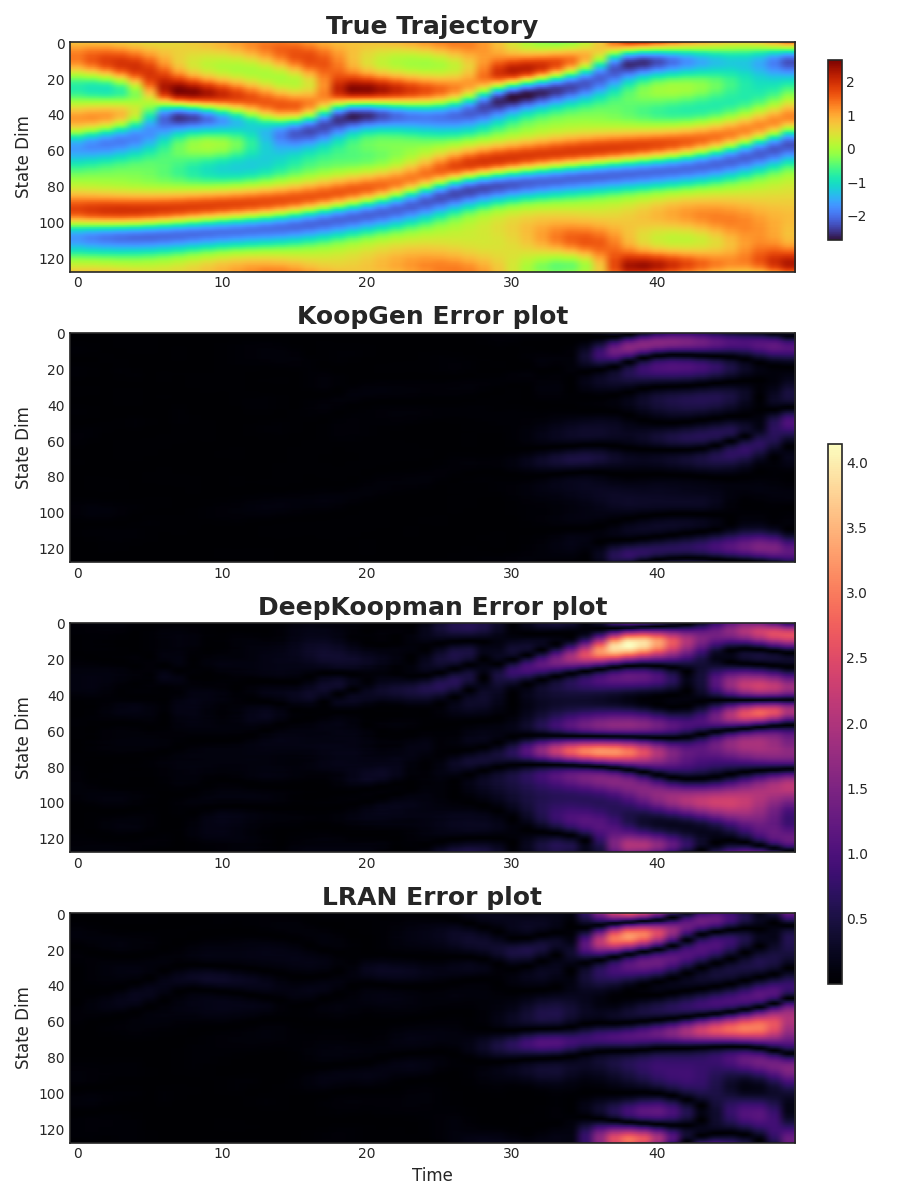}}
    \caption{Comparison of prediction errors for KoopGen, DeepKoopman and LRAN on the Kuramoto–Sivashinsky equation. While all models achieve accurate short-term predictions, KoopGen significantly suppresses long-term error growth and preserves the underlying spatiotemporal patterns, in contrast to the rapid degradation observed for DeepKoopman and LRAN.}
    \label{fig:ks_error_plot}
\end{figure}
Finally, we test KoopGen on the Kuramoto–Sivashinsky equation, a canonical model of spatiotemporal chaos,
\begin{equation}
    u_t+u_{xx}+u_{xxxx}+uu_x=0,\ \ \ \ \  x\in[0,L],
\end{equation}
on a domain $L=8\pi$, which corresponds to a regime just beyond the onset of chaos. This system presents a particularly stringent test due to the coupling of temporal chaos with broadband spatial interactions.

Some example of ground truth and corresponding predicted error on the test data sequences are shown in Fig.~\ref{fig:ks_error_plot} with additional examples provided in the Appendix~\ref{apd:res}. All methods achieve quantitatively accurate predictions over short horizons, indicating comparable expressive power in capturing the local dynamics. However, as the prediction horizon increases, their behaviors diverge markedly. Beyond approximately 30 time steps, errors in the DeepKoopman and LRAN model grow rapidly and accumulate across space, leading to visible distortion of the spatiotemporal patterns. In contrast, KoopGen maintains coherent structures and substantially lower error growth, yielding predictions that remain faithful to the underlying dynamics over extended horizons. This pronounced gap highlights the improved generalization and long-term stability of KoopGen for spatiotemporal systems.

\section{Conclusion}

We have introduced a mathematically grounded and physically interpretable neural architecture that embeds the Cartesian decomposition directly into the structure of Koopman generators, enabling a scalable and adaptive representation of dynamical systems with continuous spectra. The proposed framework extends Koopman-based learning beyond fixed or finite-dimensional operator assumptions and naturally scales to high-dimensional and spatiotemporally chaotic regimes that have traditionally been inaccessible to spectral operator methods.

Across all evaluated systems, the model achieves stable long-horizon predictions while preserving operator-theoretic consistency. Beyond predictive accuracy, the framework recovers intrinsic dynamical structure in an interpretable manner. In the Lorenz system, for instance, learned eigenfunctions faithfully reconstruct the stretching-folding geometry and wing switching dynamic of the attractor underpin chaotic behavior. This level of structural fidelity distinguishes the proposed approach from prior data-driven models and helps explain its robustness in long-term forecasting: it captures the mechanisms that generate chaos, rather than merely tracking short-term trajectories.

Despite these advances, several challenges remain. From a modeling perspective, further improvements in computational efficiency and parameter optimization are required to extend the framework to even higher-dimensional systems and longer prediction horizons. From a theoretical standpoint, a more comprehensive analysis of approximation error, stability guarantees, and generalization in the presence of finite data and noise remains an open problem. In addition, while the present study demonstrates strong performance on canonical benchmark systems, the effectiveness of the proposed approach on real-world data will be left for future study.

Overall, this work establishes a principled foundation for data-driven Koopman operator learning in continuous-spectrum systems and opens new avenues for combining operator-theoretic structure with modern neural architectures. We anticipate that continued theoretical developments and real-world validations will further broaden its applicability and deepen its impact across scientific and engineering domains.

\section{Methods}
\subsection{Generating the datasets}

We generated datasets by numerically integrating the governing differential equations in Python. For each dynamical system, 20,000 trajectories were obtained from randomly sampled initial conditions. The data were partitioned into training ($70\%$), validation ($10\%$), and testing ($20\%$) sets.

The pendulum dataset is created from random initial conditions x, where $x_1\in[-3.1,3.1]$ (just under $[-\pi,\pi]$), $x_2\in[-2,2]$, and the potential function is under 0.99 as same as the DeepKoopman \cite{lusch2018deep}. The potential function for the pendulum is $\frac 12 x_2-\cos(x_1)$. These ranges are chosen to sample the pendulum in the full phase space where the pendulum approaches having an infinite period.

The Lorenz-63 dataset is created from random initial conditions $\mathbf x=(x_1,x_2,x_3)$, where $x_1\in[-18,18]$, $x_2\in[-20,20]$, and $x_3\in[0,50]$. To ensure trajectories lie on the attractor, each trajectory is initialized from a uniformly sampled state within the above ranges and integrated forward for a burn-in time of 10.0 units (with 1000 intermediate steps) whose states are discarded. The final state of the burn-in is then used as the effective initial condition. From each such state, trajectories are simulated over $t\in[0,4]$ with step size $\Delta t=0.01$, giving $401$ snapshots per trajectory.

The Lorenz-96 dataset is created from random initial conditions independently and uniformly from the interval $[-5,5]^K$ and we take $K = 36$ variables and the classical forcing parameter $F = 8$, a regime known to produce sustained high-dimensional chaotic dynamics. Each initial condition undergoes a burn-in integration of $10$ time units to ensure that trajectories settle onto the attractor. We then simulate the system over the time interval $t \in [0,4]$ with a sampling step of $\Delta t=0.01$, producing $401$ snapshots per trajectory.

The KS dataset is created by simulating the PDE on a periodic spatial domain $x\in[0,L]$ with length $L=8\pi$. The solution is discretized with $N=128$ equispaced grid points, and advanced in time using an ETDRK4 \cite{kassam2005fourth} scheme with step size $\Delta t=1.0$. In spectral space, the linear operator is given by $\hat L(k)=k^2-k^4$, and nonlinear terms are computed via FFTs with 2/3 dealiasing to avoid aliasing errors. Initial conditions are constructed in Fourier space by exciting a small number of low-frequency modes (e.g. $m=1,2,3$) with random complex amplitudes and their conjugates, ensuring real-valued solutions. Each trajectory undergoes a spin-up phase of 20 steps, during which the transient dynamics are discarded to ensure the system reaches its statistically steady regime.

\subsection{Loss function}

Our training objective consists of two terms that jointly enforce accurate state-space prediction and consistency of the latent linear dynamics. Let $\mathbf{x}(t)$ denote the ground-truth trajectory and $\hat{\mathbf{x}}(t)$ the corresponding model prediction. The encoded Koopman eigenfunction is given by
$\mathbf{z}(t) = \Phi\bigl(\mathbf{x}(t)\bigr)$, and the latent state evolves according to the learned linear update 
\begin{equation}
    \hat{\mathbf{z}}(t)
= K\!\left(\hat{\mathbf{z}}(t-1)\right)\hat{\mathbf{z}}(t-1), \qquad \mathbf {\hat z}(0)=\mathbf z(0).
\end{equation}
To robustly capture dynamical structures, we measure errors using Sobolev norms~\cite{adams2003sobolev} rather than standard $\ell_p$ norms. Unlike pointwise losses, Sobolev norms penalize discrepancies not only in amplitude but also in temporal derivatives, thereby aligning the learning objective with the smoothness and spectral characteristics of dynamical trajectories. The overall loss function is defined as
\begin{equation}
\label{eq:loss}
\mathcal{L}
= \bigl\|\mathbf{x}(t)-\hat{\mathbf{x}}(t)\bigr\|_{k,p}+\
\alpha\bigl\|\mathbf{z}(t)-\hat{\mathbf{z}}(t)\bigr\|_{k,p},
\end{equation}
where $\alpha>0$ balances the reconstruction and linear latent dynamics consistency terms that listed in Table~\ref{tab:opt_hyperparams}, and $\|\cdot\|_{k,p}$ denotes the Sobolev norm
\begin{equation}
    \|u(t)\|_{k,p}
= \left( \sum_{i=0}^{k} \bigl\|u^{(i)}(t)\bigr\|_{p}^{p} \right)^{1/p},
\end{equation}
with $u^{(i)}(t)$ denoting the $i$-th temporal derivative of $u(t)$. 

In this work, we adopt the case $p=2$, for which the Sobolev norm~\cite{adams2003sobolev} admits a convenient Fourier-domain representation,
\begin{equation}
\label{eq:sobolev-fourier}
\|u(t)\|_{k,2}^{2}
=
\sum_{\xi=-\infty}^{\infty}
\left(1+\xi^{2}+\cdots+\xi^{2k}\right)\,\bigl|\hat{u}(\xi)\bigr|^{2},
\end{equation}
where $\hat u(\xi)$ is the Fourier transform of $u(t)$.

\subsection{Training and testing}
All models were implemented in PyTorch \cite{paszke2019pytorch} and trained on an NVIDIA RTX 3090 GPU. Network architecture parameters are shown in Table~\ref{tab:architecture}, the hyperparameter of the autoencoders, training epochs and batch size for all models are the same. During training, each trajectory $\mathbf x(t)$ is uniformly sampled at 30 equally spaced time instances and during the test, start from a randomly selected initial point set and predict 50 steps forward. Network parameters were initialized using the default PyTorch initialization scheme and optimized with the AdamW optimizer~\cite{loshchilov2018decoupled}. We adopted a partitioned optimization strategy in which the main network, gating network, and generator modules were assigned separate learning rates and training hyperparameters, as summarized in Table~\ref{tab:opt_hyperparams}. Learning rates were scheduled using a OneCycle policy~\cite{smith2019super} with a warm-up ratio of 0.1.

\begin{table}[t]
    \centering
    \caption{Network architecture for different dynamical systems}
    \begin{tabular}{lcccc}
        \toprule
        \textbf{Component} 
        & \textbf{Pendulum} 
        & \textbf{Lorenz-63} 
        & \textbf{Lorenz-96} 
        & \textbf{KS equation} \\
        \midrule
        Main network depth 
        & 3 & 4 & 4 & 4 \\
        Main network width 
        & 32 & 128 & 256 & 1024 \\
        Latent dimension 
        & 2 & 6 & 64 & 256 \\
        Gating network depth 
        & 2 & 2 & 2 & 2 \\
        Gating network width 
        & 32 & 128 & 256 & 1024 \\
        Number of $\widehat{G}$
        & 2 & 6 & 64 & 32 \\
        Number of  $\widetilde{G}$
        & 0 & 2 & 8 & 16 \\
        Step $\Delta t$&0.1&0.1&0.1&0.1\\
        \bottomrule
    \end{tabular}
    \label{tab:architecture}
\end{table}

\begin{table}[t]
    \centering
    \caption{Optimization hyperparameters}
    \begin{tabular}{lcccc}
        \toprule
        \textbf{Hyperparameter}
        & \textbf{Pendulum}
        & \textbf{Lorenz--63}
        & \textbf{Lorenz--96}
        & \textbf{KS} \\
        \midrule
        LR (Main Net)      
        & 0.005 & 0.005 & 0.005 & 0.001 \\
        LR (Gate Net)      
        & 0.001 & 0.001 & 0.001 & 0.001 \\
        LR (Generators)    
        & 0.005 & 0.005 & 0.005 & 0.001 \\
        Loss weight $\alpha$
        & 0.1 & 1.0 & 0.1 & 1.0 \\
        Training epochs             
        & 100 & 200 & 500 & 500 \\
        Batch Size   
        & 128 & 128 & 128 & 128 \\
        \bottomrule
    \end{tabular}
    \label{tab:opt_hyperparams}
\end{table}



\bibliography{sn-bibliography}

\begin{appendices}

\section{Experiment settings}
The hyperparameters used for DeepKoopman and LRAN are summarized in Tables~\ref{tab:DK} and~\ref{tab:LRAN}, respectively. Unless otherwise specified, all remaining architectural and optimization settings are identical to those used for KoopGen and are reported in Tables~\ref{tab:architecture} and~\ref{tab:opt_hyperparams}.

\begin{table}[tb]
\centering
\caption{Hyperparameter settings for DeepKoopman}
    \begin{tabular}{lcccc}
    \toprule
    \textbf{Hyperparameter}
        & \textbf{Pendulum}
        & \textbf{Lorenz-63}
        & \textbf{Lorenz-96}
        & \textbf{KS} \\
        \midrule
        Learning rate & 0.001& 0.001 &0.001&0.0001\\
        Num complex eigenvalues & 1& 3 &32&512\\
        Num real eigenvalues & 0& 0 &0&0\\
        Loss weight $\alpha$ & 0.1& 1.0 &0.1 &1.0\\
         \bottomrule
    \end{tabular}
    \label{tab:DK}
\end{table}

\begin{table}[tb]
    \centering
    \caption{Hyperparameter settings for LRAN}
    \begin{tabular}{lcccc}
    \toprule
    \textbf{Hyperparameter}
        & \textbf{Pendulum}
        & \textbf{Lorenz-63}
        & \textbf{Lorenz-96}
        & \textbf{KS} \\ \midrule
        Learning rate & 0.001& 0.001 &0.001&0.001\\
        Loss weight $\alpha$ & 0.1& 1.0 &0.1 &0.1\\
     \bottomrule
    \end{tabular}
    \label{tab:LRAN}
\end{table}

\section{More results}\label{apd:res}
\begin{figure}[htb]
    \centering
    \includegraphics[width=0.7\linewidth]{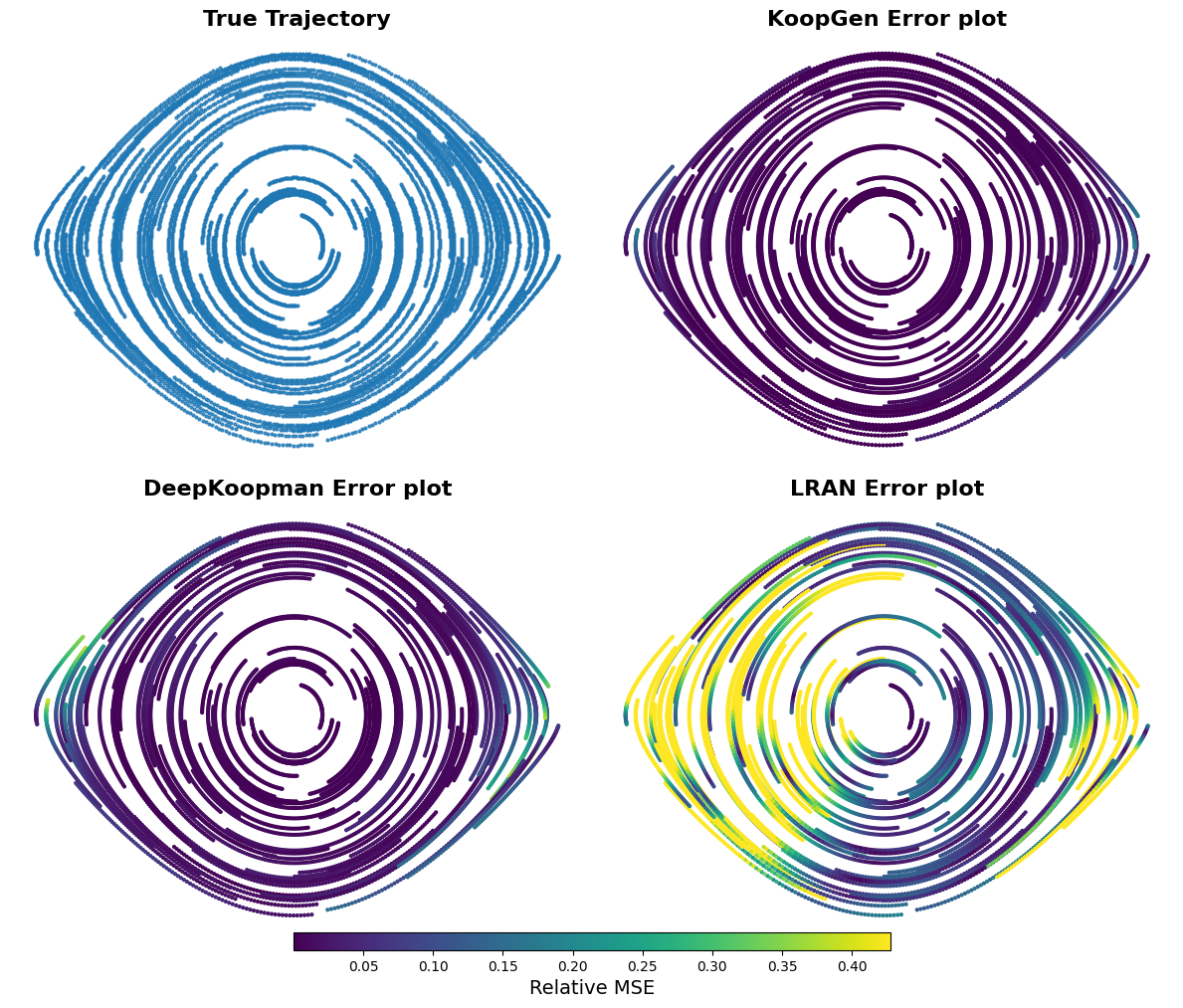}
    \caption{Comparison of prediction errors for KoopGen, DeepKoopman and LRAN on Pendulum systems.}
    \label{fig:pen_error}
\end{figure}

\begin{figure}[htb]
    \centering
    \includegraphics[width=0.8\linewidth]{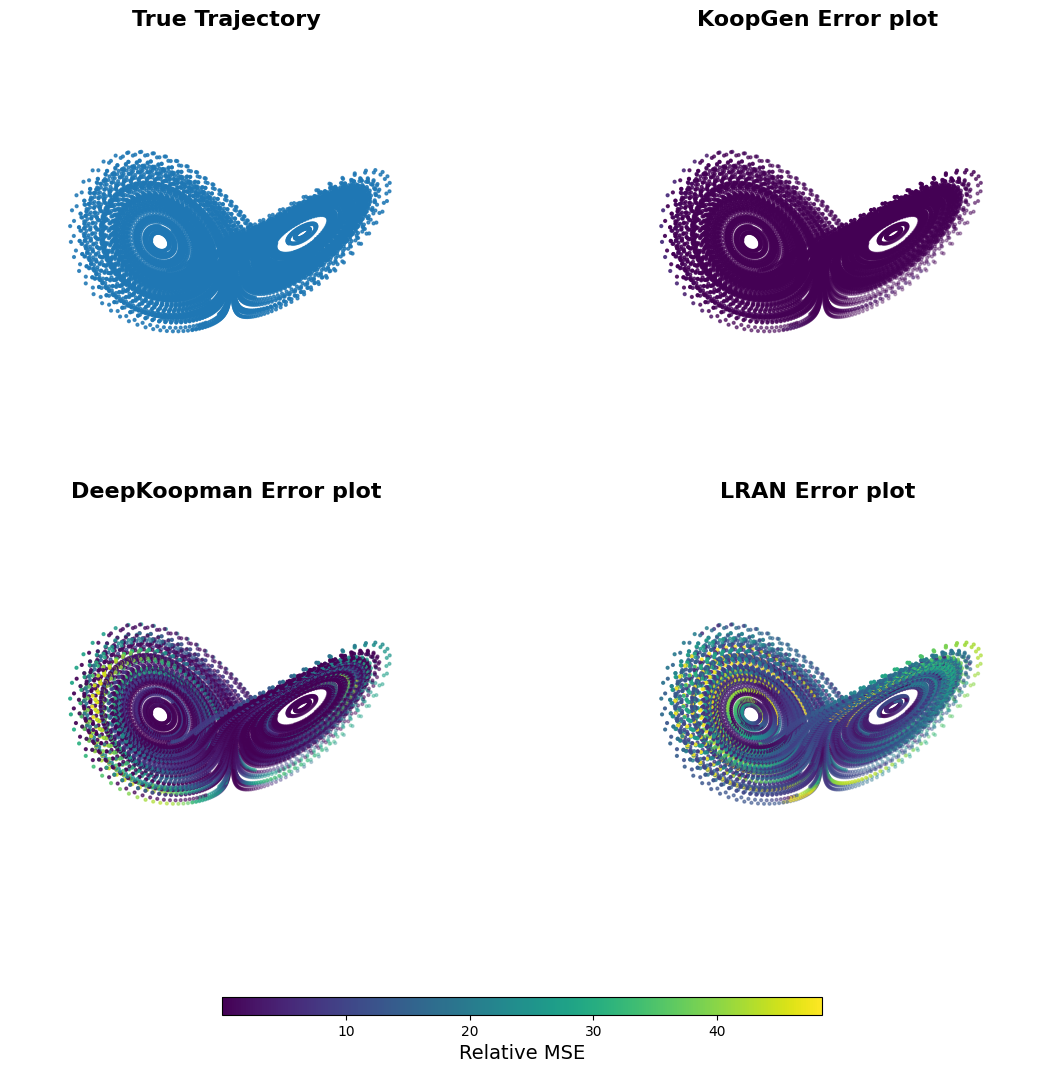}
    \caption{Comparison of prediction errors for KoopGen, DeepKoopman and LRAN on Lorenz-63 systems.}
    \label{fig:lor_error}
\end{figure}

\begin{figure}[tb]
    \centering
    \subfloat[\label{fig:96_0}]{\includegraphics[width=0.45\linewidth]{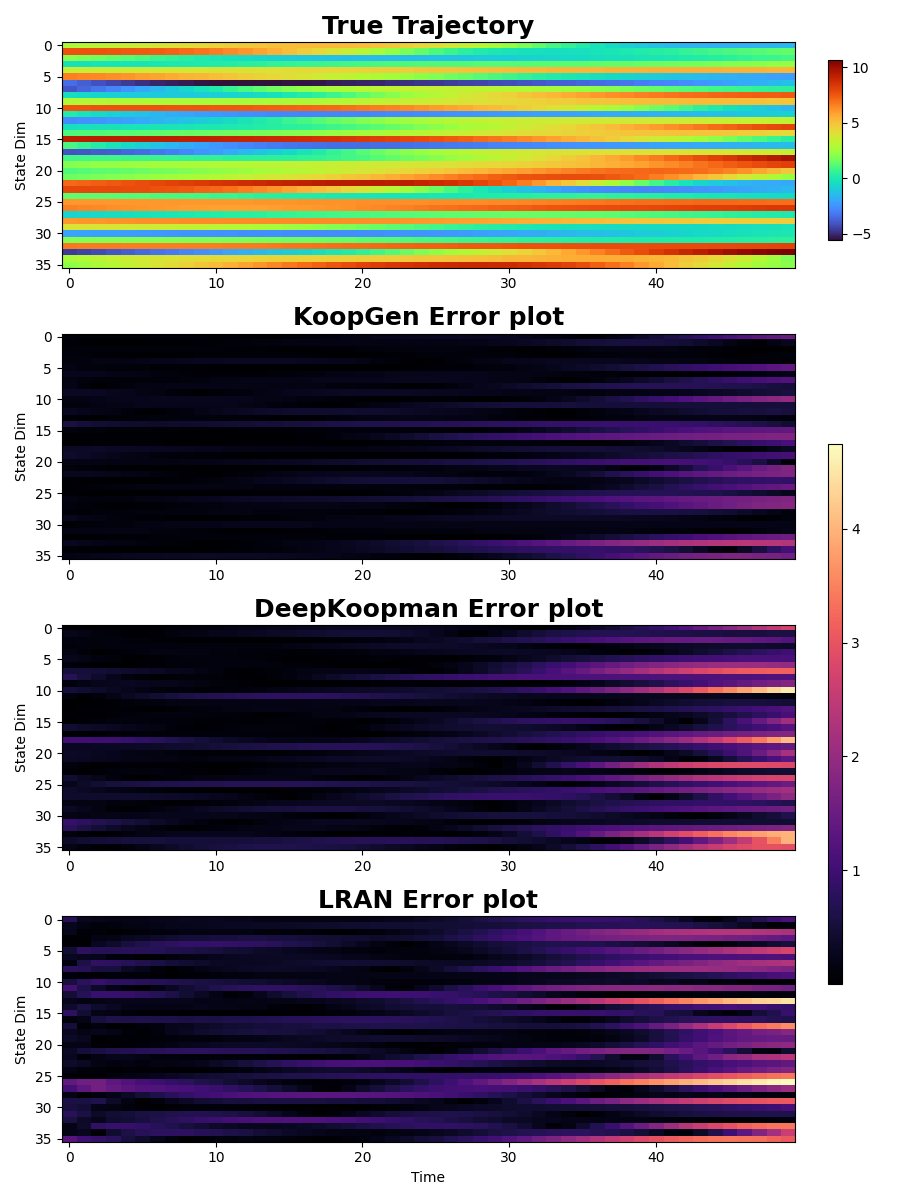}}
    \hfill
    \subfloat[\label{fig:96_14}]{\includegraphics[width=0.45\linewidth]{figures/lorenz96_14th_batch.png}}
    \hfill
    \subfloat[\label{fig:96_23}]{\includegraphics[width=0.45\linewidth]{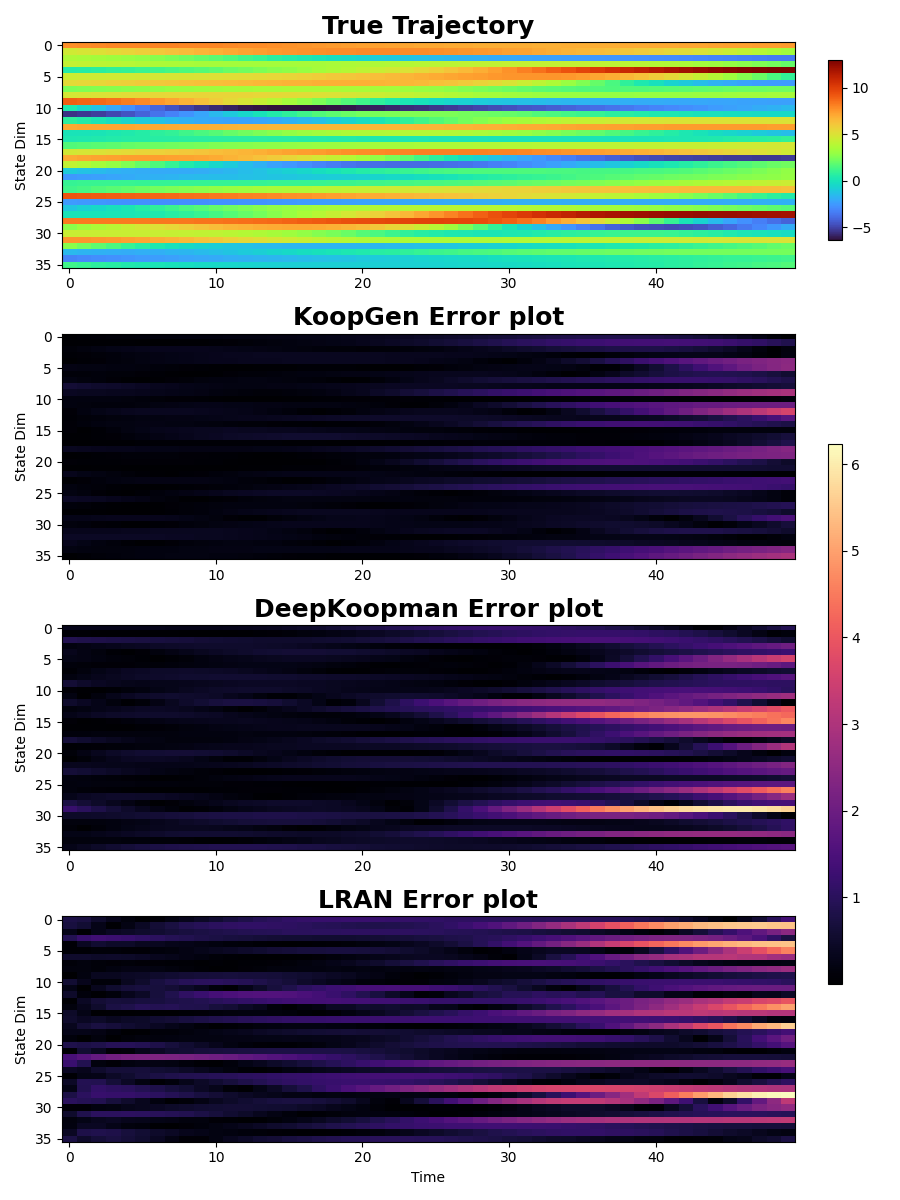}}
    \hfill
    \subfloat[\label{fig:96_31}]{\includegraphics[width=0.45\linewidth]{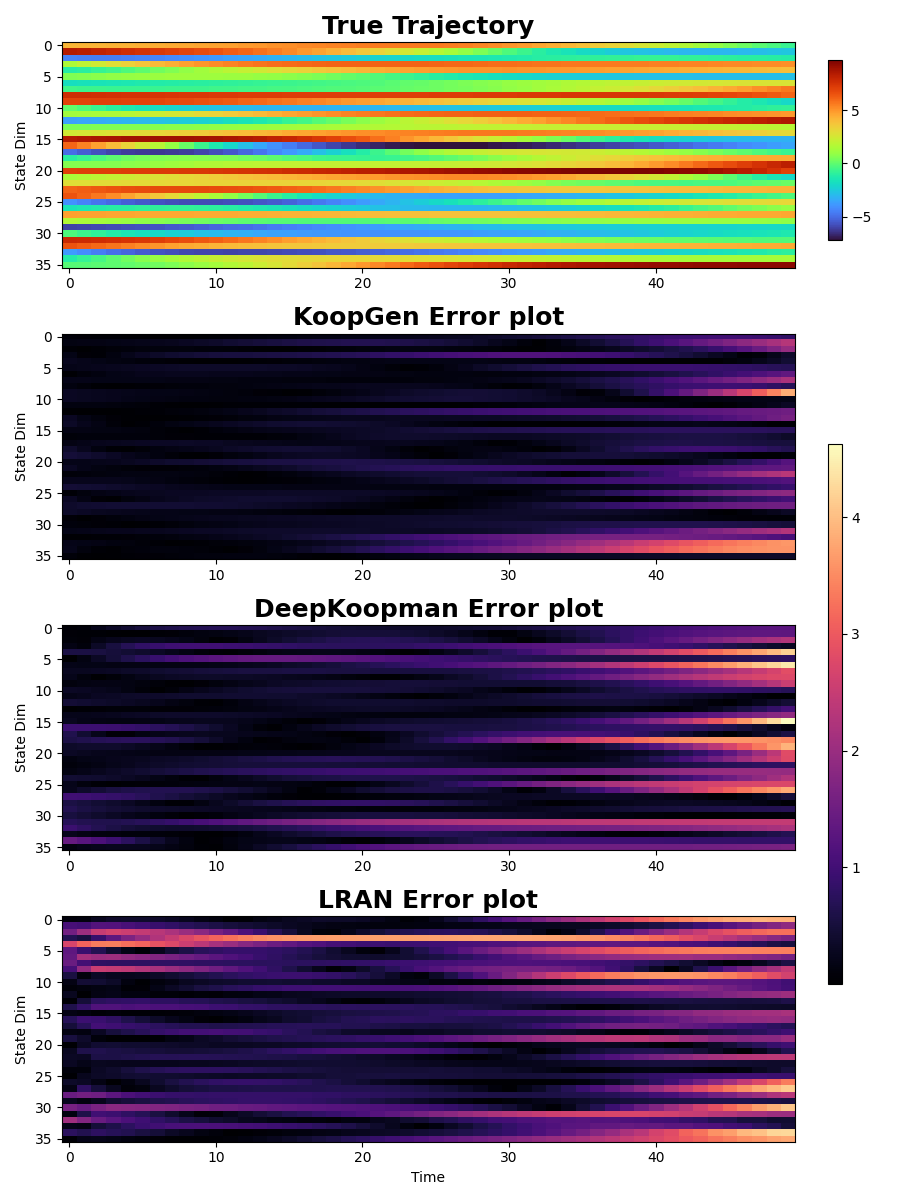}}
    \caption{Comparison of prediction errors for KoopGen, DeepKoopman and LRAN on Lorenz-96 system.}
    \label{fig:96_error}
\end{figure}

\begin{figure}[tb]
    \centering
    \subfloat[\label{fig:KS_3}]{\includegraphics[width=0.45\linewidth]{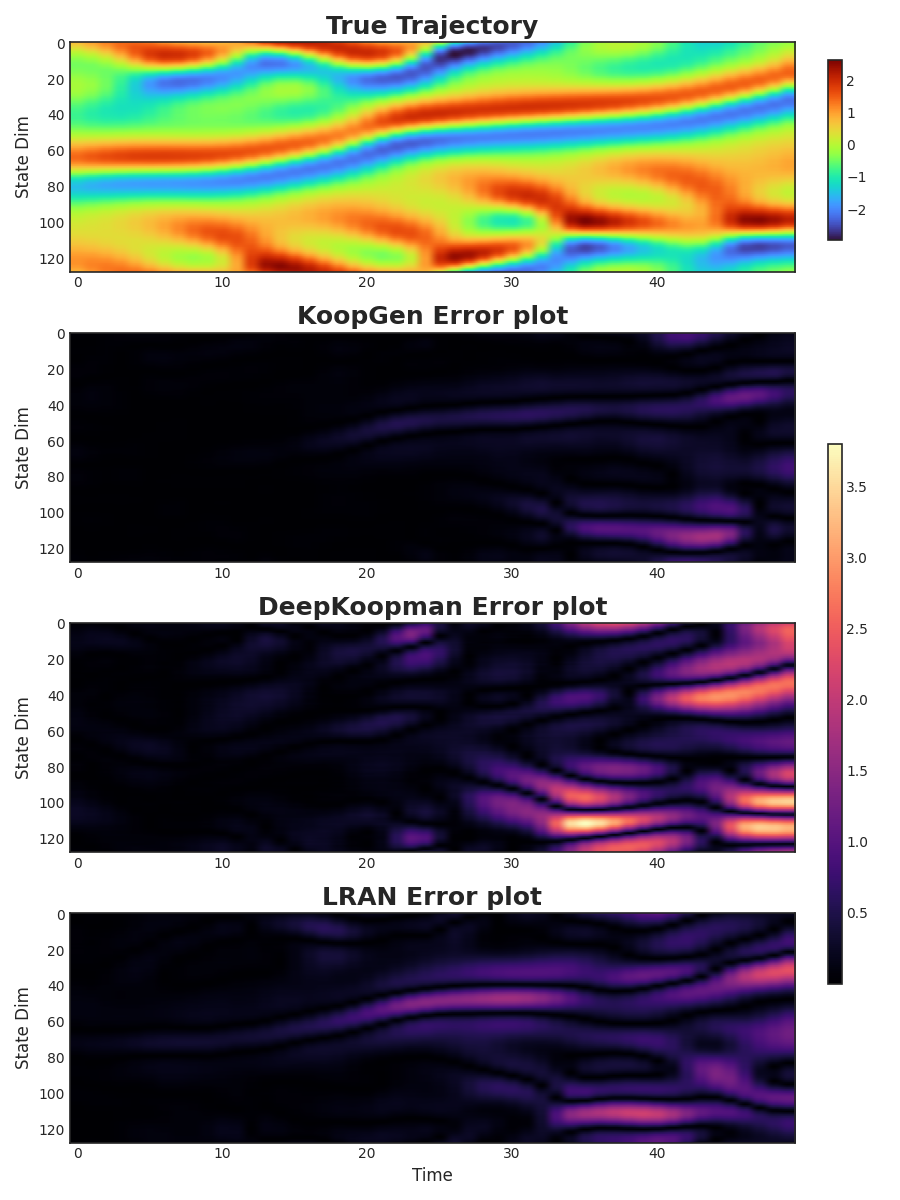}}
    \hfill
    \subfloat[\label{fig:KS_10}]{\includegraphics[width=0.45\linewidth]{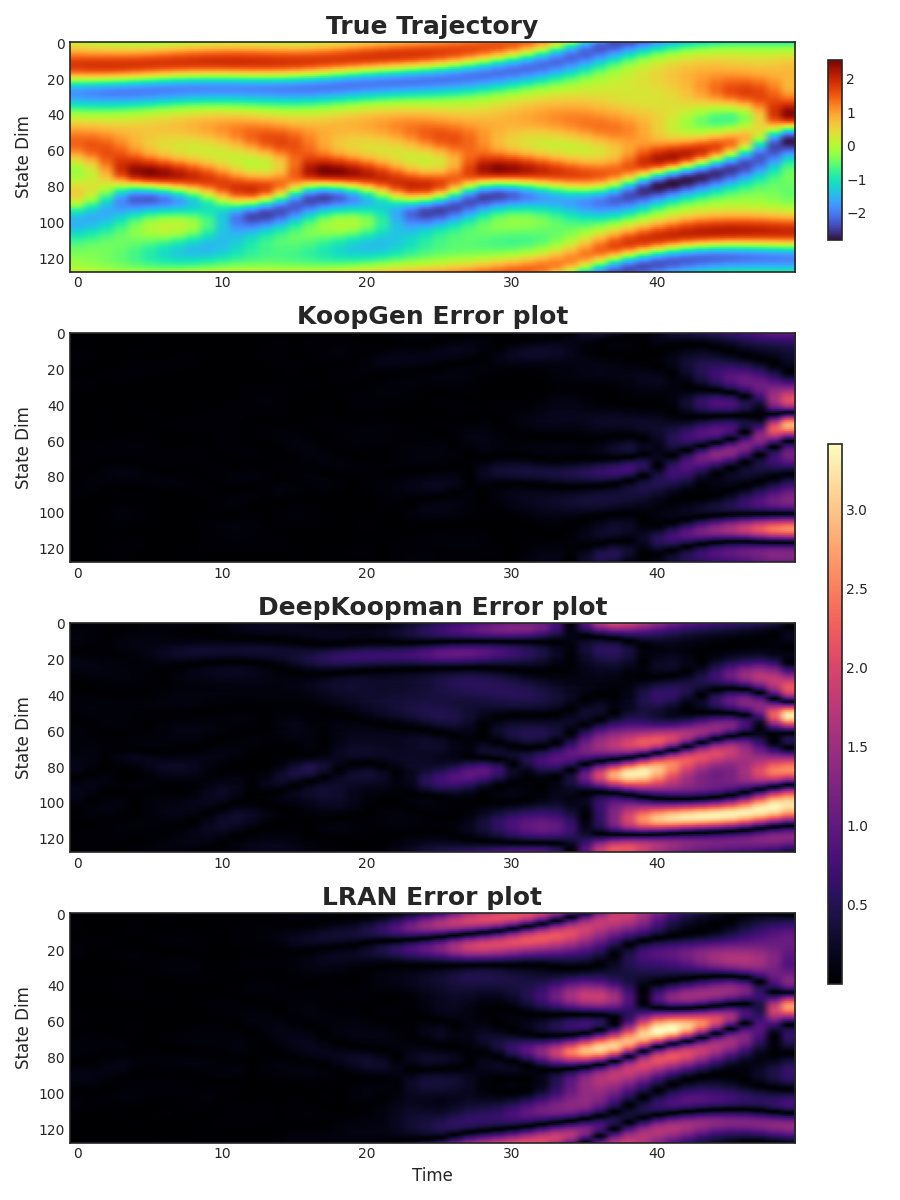}}
    \hfill
    \subfloat[\label{fig:KS_23}]{\includegraphics[width=0.45\linewidth]{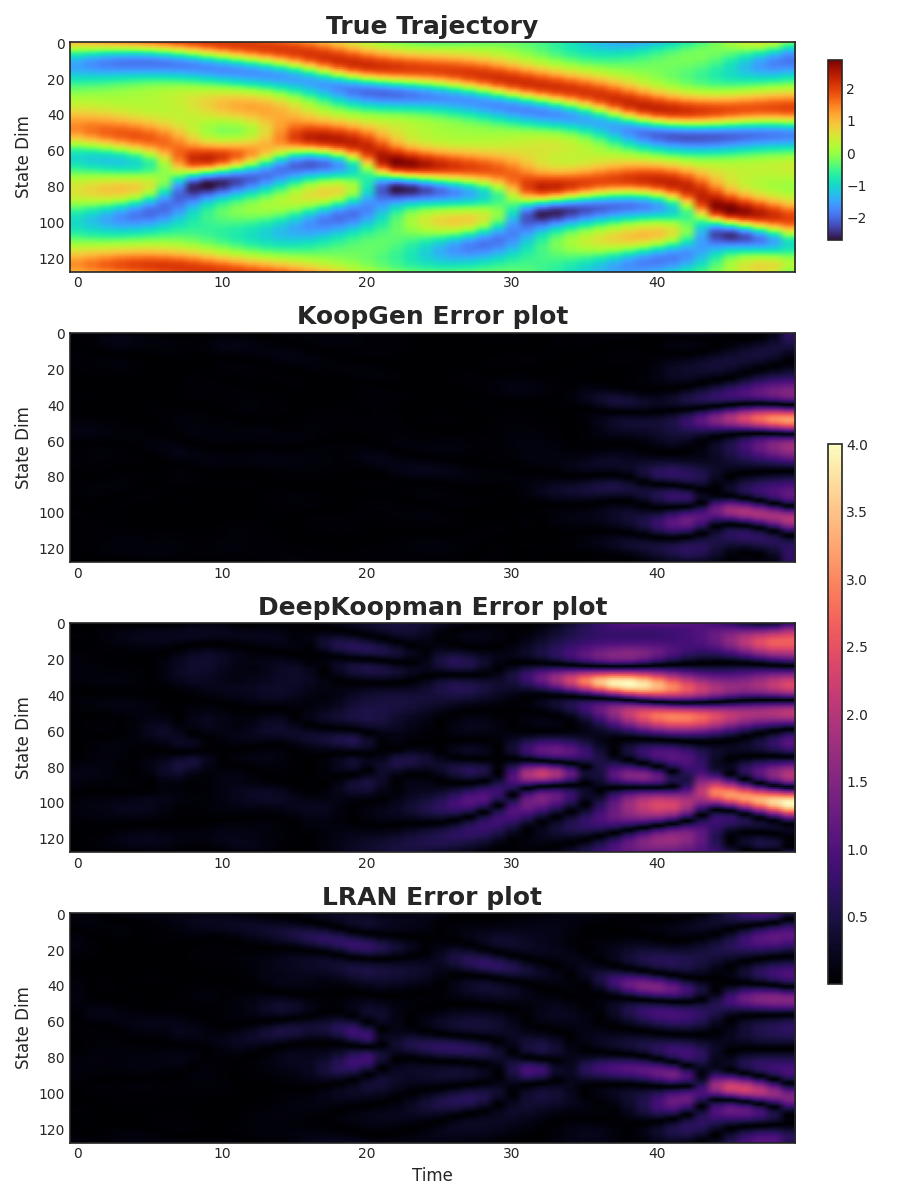}}
    \hfill
    \subfloat[\label{fig:KS_28}]{\includegraphics[width=0.45\linewidth]{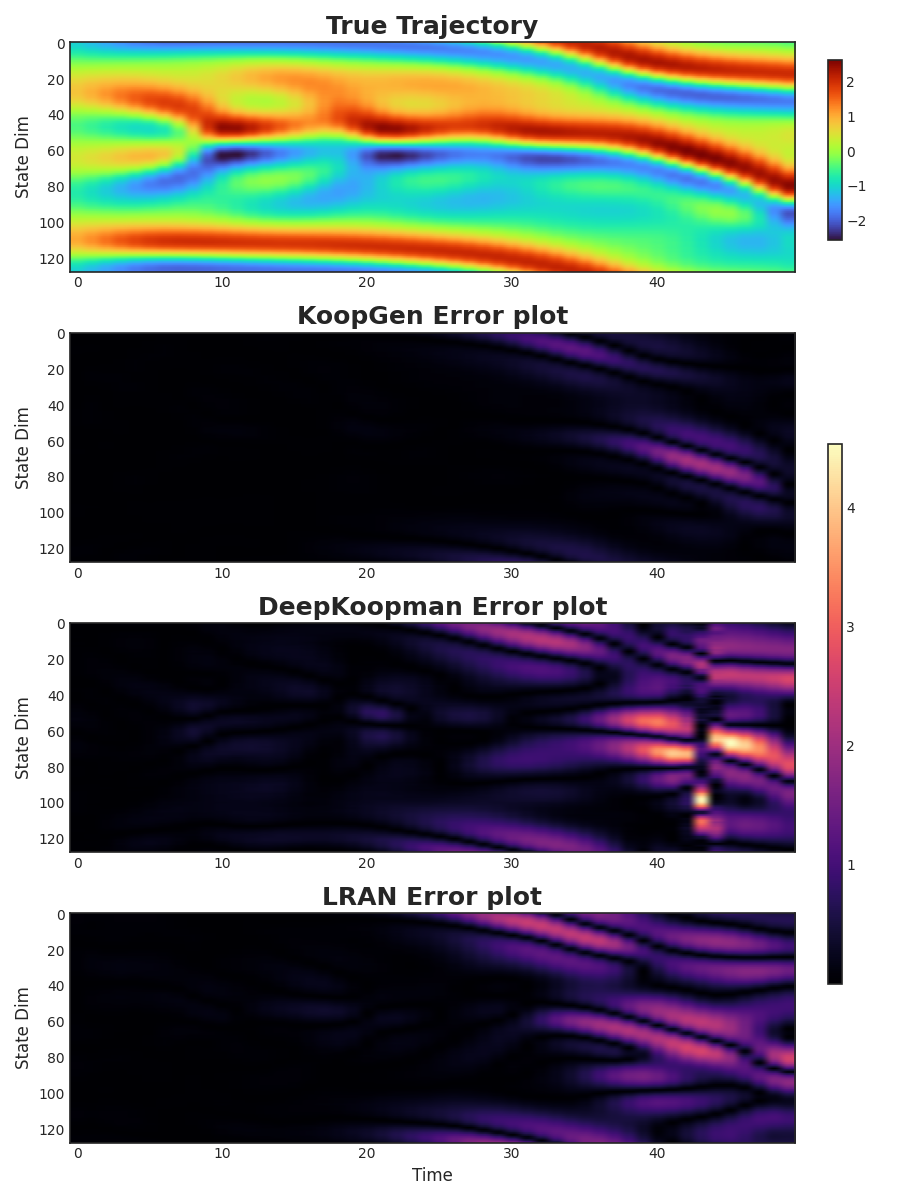}}
    \caption{Comparison of prediction errors for KoopGen, DeepKoopman and LRAN on KS equation.}
    \label{fig:ks_error}
\end{figure}
This section provides additional prediction results for all methods and systems considered in the main text, serving as a supplementary and more detailed illustration of their behaviors under a wider range of initial conditions. Overall, the trends observed here are consistent with those reported in the main text. All error plots report the absolute prediction error, $\vert \hat x -x\vert $.

Fig.~\ref{fig:pen_error} and Fig.~\ref{fig:lor_error} present the prediction error for nonlinear Pendulum and Lorenz-63 system, respectively. For each system, we evaluate 50-step forward predictions from 4,000 randomly sampled initial conditions. For these low-dimensional continuous-spectrum systems, state-dependent Koopman formulations yield markedly improved predictions compared with state-independent operators, highlighting the importance of adaptive transfer operators. In particular, KoopGen produces error distributions that remain relatively uniform across the attractor, indicating stable predictive behavior over a broad range of states. As shown in Fig.~\ref{fig:lor_error}, DeepKoopman produces pronounced error amplification near the outer regions of the attractor’s two lobes, where trajectories exhibit strong sensitivity to initial conditions. These localized error bursts indicate limited robustness in capturing the global geometry and instability structure of chaotic dynamics.

Fig.~\ref{fig:96_error} and Fig.~\ref{fig:ks_error} further present prediction error distributions for the high-dimensional Lorenz–96 system and the Kuramoto–Sivashinsky equation. These systems pose substantially greater challenges due to their high dimensionality, broadband spectra, and strong spatiotemporal chaos. KoopGen maintains comparatively low prediction errors and exhibits more coherent long-term evolution over multiple prediction steps. This behavior suggests that the generator-based parameterization enables the model to better accommodate the complex interplay between Existing deep Koopman and neural  and dissipation that governs high-dimensional chaotic dynamics.

Taken together, these supplementary results provide a more detailed view of the predictive characteristics of KoopGen across a diverse set of dynamical regimes. They further illustrate how incorporating structural information at the network architecture can support stable and scalable learning of continuous-spectrum systems.






\end{appendices}

\end{document}